\documentclass[sigplan,10pt,screen]{acmart}
\renewcommand\footnotetextcopyrightpermission[1]{}
\acmConference{EuroSys '23}{May 08--12, 2023}{Rome, Italy}

\usepackage{pythonhighlight}
\usepackage{amsmath}
\usepackage{amsfonts}
\usepackage[ruled, lined, linesnumbered, nofillcomment, longend]{algorithm2e}
\usepackage[english]{babel}
\usepackage{tikz}
\usepackage{booktabs}
\usepackage{graphicx}
\usepackage[export]{adjustbox}
\usepackage{pdfpages}
\usepackage{varwidth}
\usepackage{subcaption}
\usepackage{xcolor}
\usepackage{xspace}
\usepackage{multirow}
\usepackage{hyperref}
\usepackage[inline,shortlabels]{enumitem}
\usepackage{makecell}
\usepackage{tabularx}
\usepackage{microtype}
\usepackage{booktabs}
\usepackage{float}
\usepackage{appendix}

\newcommand{\name}{\textsc{Egeria}\xspace}
\newcommand{\sys}{\textsc{Egeria}\xspace}

\graphicspath{{figures/}{../figures/}}
\DeclareGraphicsExtensions{.png,.pdf}

\begin{document}

\title{\sys: Efficient DNN Training with Knowledge-Guided Layer Freezing}

\author{
    {
        Yiding Wang$^{1}$, 
        Decang Sun$^{1}$, 
        Kai Chen$^{1}$, 
        Fan Lai$^{2}$, 
        Mosharaf Chowdhury$^{2}$
    }
    \smallskip
    \\
    {
        $^{1}$iSING Lab, Hong Kong University of Science and Technology\\
        $^{2}$University of Michigan
    }
}

\begin{abstract}

Training deep neural networks (DNNs) is time-consuming. While most existing solutions try to overlap/schedule computation and communication for efficient training, this paper goes one step further by {\em skipping} computing and communication through DNN layer freezing. 
Our key insight is that the training progress of internal DNN layers differs significantly, and front layers often become well-trained much earlier than deep layers.
To explore this, we first introduce the notion of \emph{training plasticity} to quantify the training progress of internal DNN layers. 
Then we design \name, a knowledge-guided DNN training system that employs semantic knowledge from a reference model to accurately evaluate individual layers' training plasticity and safely freeze the converged ones, saving their corresponding backward computation and communication. Our reference model is generated on the fly using quantization techniques and runs forward operations asynchronously on available CPUs to minimize the overhead.
In addition, \name caches the intermediate outputs of the frozen layers with prefetching to further skip the forward computation. 
Our implementation and testbed experiments with popular vision and language models show that \name achieves 19\%-43\% training speedup w.r.t. the state-of-the-art without sacrificing accuracy.
\end{abstract}

\maketitle
\renewcommand{\shortauthors}{Wang et al.}

\section{Introduction}
\label{sec:intro}

Recent advances in deep learning (DL) benefit significantly from training larger deep neural networks (DNNs) on larger datasets.
Due to growing model sizes and large volumes of data, DNNs have become more computationally expensive to train, raising the cost and carbon emission of large-scale training~\cite{patterson2021carbon}.
Many recent DL research works focus on improving parallelism and pipelining via sophisticated computation-communication overlapping or scheduling to build more efficient systems and reduce training time~\cite{narayanan2019pipedream,peng2019generic,jiang2020unified}.
Nevertheless, while approaching linear scalability can reduce the time to train a model, the total amount of computation requirement remains the same.

In this paper, we move one step further to explore: {\em can we reduce the total computation (and communication) in large DNN training?}
We propose a \emph{knowledge-guided} training system, \name, to accelerate DNN training via computation-communication freezing while still maintaining accuracy. Our key insight is that the training progress of internal DNN layers differs significantly, and front layers can become well-trained much earlier than deep layers. 
This is because DNN features transition from being task-agnostic to task-specific from the first to the last layer~\cite{NIPS2014_5347,zeiler2014visualizing}. Thus, the front layers of a DNN often converge quickly, while the deep layers take a much longer time to train, as generally observed in both vision and language models~\cite{shen2020reservoir,rogers2020primer}, experimentally validated in \S\ref{subsec:oppo}.
\sys can safely freeze these converged DNN layers earlier, saving their corresponding computation and communication expenses without hurting model accuracy.

While freezing layers can reduce training cost, prematurely freezing under-trained layers will hurt the final accuracy.
We observe that in transfer learning, freezing layers is mainly used for solving the overfitting problem~\cite{cs231n}.
While techniques such as static freezing~\cite{lee2019would} and cosine annealing~\cite{brock2017freezeout} can reduce backward computation cost, accuracy loss is a common side effect. Thus,
the main challenge of extending layer freezing to general DNN training is how to maintain accuracy by only freezing the converged layers.

To address this challenge, \sys introduces the notion of \emph{training plasticity}\footnote{\emph{Plasticity} quantifies a layer's training progress toward convergence, which is borrowed from \emph{neuroplasticity} in neural science and child development~\cite{costandi2016neuroplasticity}. Basically, a DNN layer's training plasticity will gradually decrease and become stable as it converges.} to quantify a layer's training progress and safely detect the converged DNN layers to avoid premature freezing.
To this end, \name uses a \emph{reference model}, which is the proxy for semantic knowledge, to evaluate DNN layers' plasticity.
The reference model, in essence, is a trained compressed DNN with the same architecture as the model being trained to understand layer-wise performance (details in \S\ref{subsec:ref_model}).
We compare the intermediate activations (internal outputs) between the training model and the reference model elicited by the same data batch to measure the plasticity.
When the plasticity becomes stationary, it implies that the layer is converged and can be frozen safely (\S\ref{subsec:measure_maturity}).
In addition, \name can unfreeze the frozen layers to continue training with learning rate decay.
Our approach is informed by recent advances in knowledge distillation research that suggest the same input data (images and word vectors) will elicit similar \emph{pair-wise} activation patterns in \emph{trained models}~\cite{aguilar2020knowledge,Tung_2019_ICCV,park2021emnlp}.
Compared to the straightforward metric of gradient's granularity or norm against a hard label, intermediate activation as soft distribution is proved to be more semantically meaningful, and thus more accurate by ML research (\S\ref{subsubsec:plasticity}).

\name adaptively generates the reference model by instantly compressing a snapshot of the training model via quantization~\cite{han2015deep} on CPUs after the \emph{bootstrapping stage}~\cite{achille2018critical,agarwal2021adaptive} (early iterations during which the training model converges quickly).
Large DNNs are robust to quantization, according to our evaluation and ML literature~\cite{li2020train}.
\name also profiles in the background to make sure the CPU-efficient reference model can provide accurate plasticity evaluation.
The reference model exploits available CPU cores during GPU-heavy training, running forward operations parallel to the GPU training using the same input data in a non-blocking and asynchronous fashion.
Hence, the system overhead is minimal and can be well hidden.
In the remaining training process, \name updates the reference model using the latest snapshots to stabilize the plasticity curve.
Essentially, we trade off small CPU resources for maintaining accuracy when freezing layers.

Freezing the front layers can save the backward computation and parameter synchronization. Nevertheless, we find that the forward pass still takes up to 35\% of the time of an iteration.
We observe that, in DNN training, the frozen front layers will produce the same forward output given the same input. Prior work on inference also shows that caching forward results can improve performance~\cite{balasubramanian2021accelerating}. To take advantage of this, \sys saves the intermediate activations of the frozen layers to the disk, \emph{prefetches} the saved activation tensors to the GPU memory, and continues training the remaining layers from the cached activations in the following epochs. As the data loader knows the future data sequence, it is possible to prefetch relevant activations without stalling.
Caching and prefetching are also compatible with random data augmentation. Therefore, we further save the frozen layers' forward computation without altering the training data sequence ($\S$\ref{subsec:freezing}).

We implement \name as a framework-independent Python library ($\S$\ref{sec:implementation}).
Existing code can work with \name with minimal changes.
We evaluate \name using seven popular vision and language models on five datasets ($\S$\ref{sec:evaluation}).
It achieves 19\%-43\% training speedup than the state-of-the-art frameworks and can reach the target accuracy.

To summarize, the key contributions of \sys include:
\begin{enumerate*}[(1)]
    \item leveraging semantic knowledge to save the backward computation and communication via DNN layer freezing while maintaining accuracy;
    \item building an efficient system to implement the idea of knowledge-guided training; and
    \item caching the intermediate results with prefetching to further save the forward pass of the frozen layers with negligible overheads.
\end{enumerate*}
\section{Background and Motivation}
\label{sec:background}

\subsection{DNN Training}
\label{subsec:background}

Modern DNNs consist of dozens or hundreds of layers that conduct mathematical operations.
Each layer takes an input tensor of features and outputs corresponding activations.
We train a DNN by iterating over a large dataset many times and minimizing a loss function.
The dataset is partitioned into \emph{mini-batches}, and a pass through the full dataset is called an \emph{epoch}.
A DNN training iteration includes three steps:
\begin{enumerate*}[(1)]
  \item forward pass,
  \item backward pass, and
  \item parameter synchronization.
\end{enumerate*}
The forward and backward passes require GPU computation.
In each iteration, the \emph{forward pass} (FP) takes a mini-batch and goes through the model layer-by-layer to calculate the loss regarding the target labels and the loss function.
In the \emph{backward pass} (BP), we calculate the parameter gradients from the last layer to the first layer based on the chain rule of derivatives regarding the loss~\cite{maclaurin2015autograd}.
At the end of each iteration, we update the model parameters with an optimization algorithm, such as stochastic gradient descent (SGD)~\cite{bottou2010large}.
In data parallel distributed training, independently computed gradients from all workers are aggregated over the network to update the shared model.

\subsection{Existing Optimizations for DNN Training}\label{subsec:comm_opt}
Training large DNNs is computation- (and communication-) intensive due to the ever-growing data volumes and model size~\cite{narayanan2019pipedream,peng2019generic,jayarajan2019priority}.
One important direction for training acceleration from the system perspective closely related to \sys is \emph{computation-communication overlapping and scheduling}.
Baseline training frameworks (e.g., TensorFlow, PyTorch, and Poseidon~\cite{zhang2017poseidon}) optimize distributed performance by issuing the gradient transmission once a layer finishes its backward computation so that the deeper layers can overlap their communication with the front layer's BP.
Priority-based communication scheduling systems (e.g., ByteScheduler~\cite{peng2019generic}, P3~\cite{jayarajan2019priority} and TicTac~\cite{hashemi2018tictac}) leverage the layer-wise structure information to prioritize the front layers in communication which try to overlap the communication with FP. 
Pipelining solutions~\cite{narayanan2019pipedream, huang2018gpipe, yang2021pipemare} add inter-batch pipelining to intra-batch parallelism to further improve parallel training throughput.
While all these solutions can optimize computation-communication efficiency, the total computational cost remains the same.

\paragraph{Other methods.}
There are other optimizations, such as gradient sparsification~\cite{lin2018deep} and quantization~\cite{wen2017terngrad}, to reduce communication volumes. 
These methods are largely orthogonal to \sys, and we will overview them in \S\ref{sec:related}. 

\subsection{Opportunities of DNN Layer Freezing}\label{subsec:oppo}
In this paper, we explore the idea of {\em reducing computation and communication costs} through DNN layer freezing. 
In the following, we first show the idea and its potential, and then lay out the challenges, motivating the design of \sys. 

\begin{figure}[!t]
  \centering
  \includegraphics[width=\columnwidth]{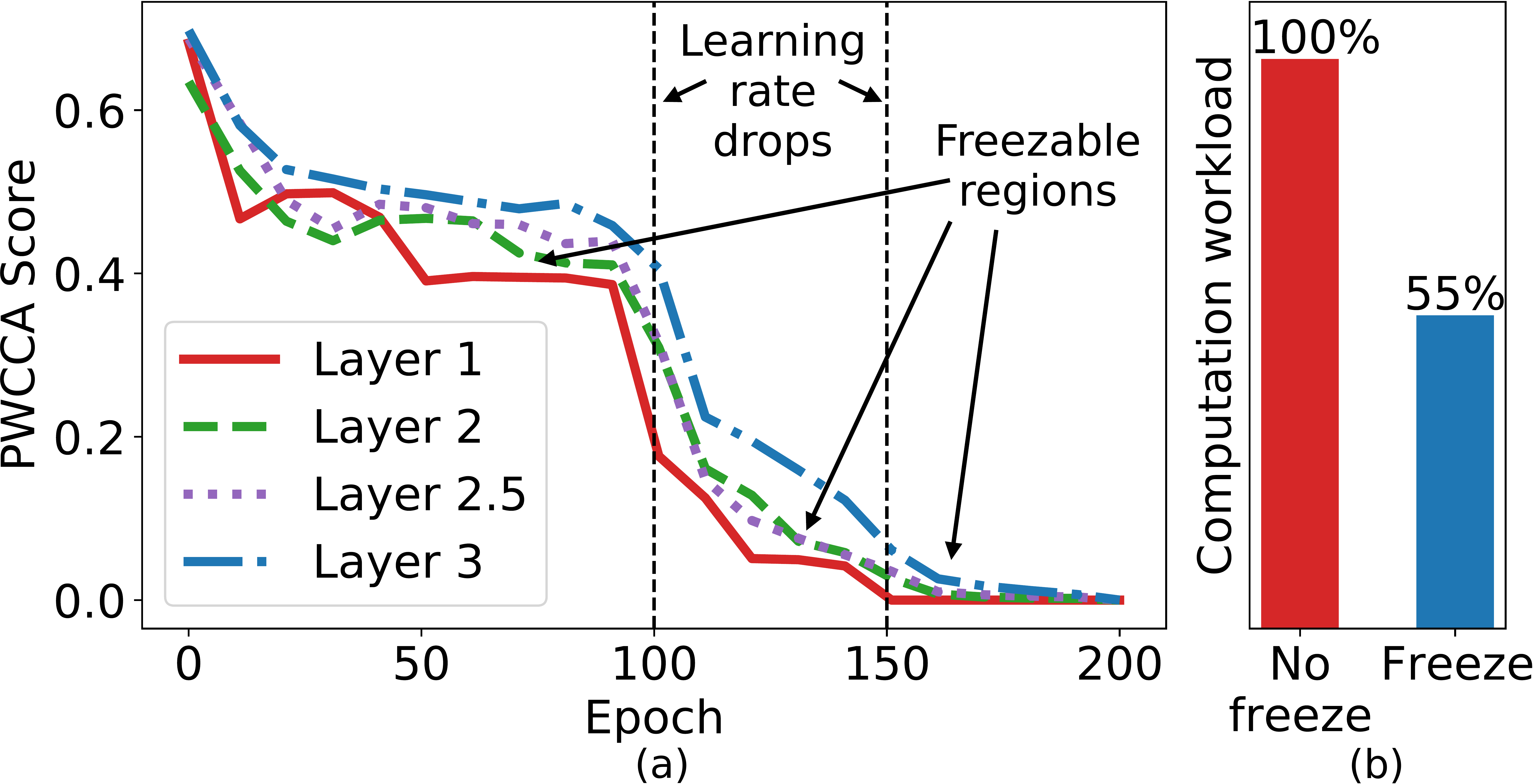}
  \caption{Post hoc layer convergence analysis with PWCCA. A lower score means the layer converges better. Layer module 2.5 is the first half of Layer module 3. The learning rate changes at 100th and 150th epochs and reboosts training. We can freeze layers when they are stationary and unfreeze them when the learning rate decreases extremely.}
  \label{fig:pwcca}
\end{figure}

\paragraph{Motivation for layer freezing.} 
Recent efforts have shown that the front layers primarily extract general features of the raw data (e.g., the shape of objects in an image) and often become well-trained much earlier, while deeper layers are more task-specific and capture complicated features output from front layers~\cite{NIPS2014_5347,raghu2017svcca}.
Our work is also inspired by transfer learning~\cite{long2015learning,howard2018universal,fine_tune_pytorch,fine_tune_keras}.
When fine-tuning a pre-trained model on a new task, we find that ML practitioners can freeze (i.e., fix layer's weights) the front layers or only fine-tune them for a few iterations and focus on training the deep layers on the new dataset.

To demonstrate the potential in general training (i.e., not fine-tuning), we use PWCCA~\cite{morcos2018insights}, a \emph{post hoc} layer convergence analysis tool, to track the training progress of different layers of ResNet-56~\cite{he2016identity} as an example.
ResNet-56 is a popular model for image classification on the CIFAR-10 dataset, and it consists of three layer blocks (refereed to as modules or stages), where each module has 18 basic blocks of successive layers.
PWCCA compares the intermediate activation (i.e., the output feature map produced by a DNN layer) with a \emph{fully-trained} model.
A low PWCCA score (0\textendash1 range) suggests the layer is converged to the final state.
We can clearly find some freezable regions in Figure~\ref{fig:pwcca}: e.g., for layer 1, during the 50th\textendash90th, 120th\textendash140th, and after the 160th epoch, its score becomes stable, meaning it is temporarily converged; other layers also show some relatively stable regions.
The scores drop at the 100th and 150th epochs because the learning rate decreases as scheduled; after that, they soon converge again.
These patterns reveal a natural strategy: \emph{Freeze the layers when their performance is stable and unfreeze them when the learning rate decreases.}

We find that if we freeze the layers in their freezable regions, by summing up the \#parameters when a layer's PWCCA is stable in Figure~\ref{fig:pwcca}, we can reduce the computation costs by 45\% in theory!
For natural language processing (NLP) models, this potential can be even larger because the front layers usually contain more parameters than CNNs.

\begin{figure}[!t]
  \includegraphics[width=\columnwidth]{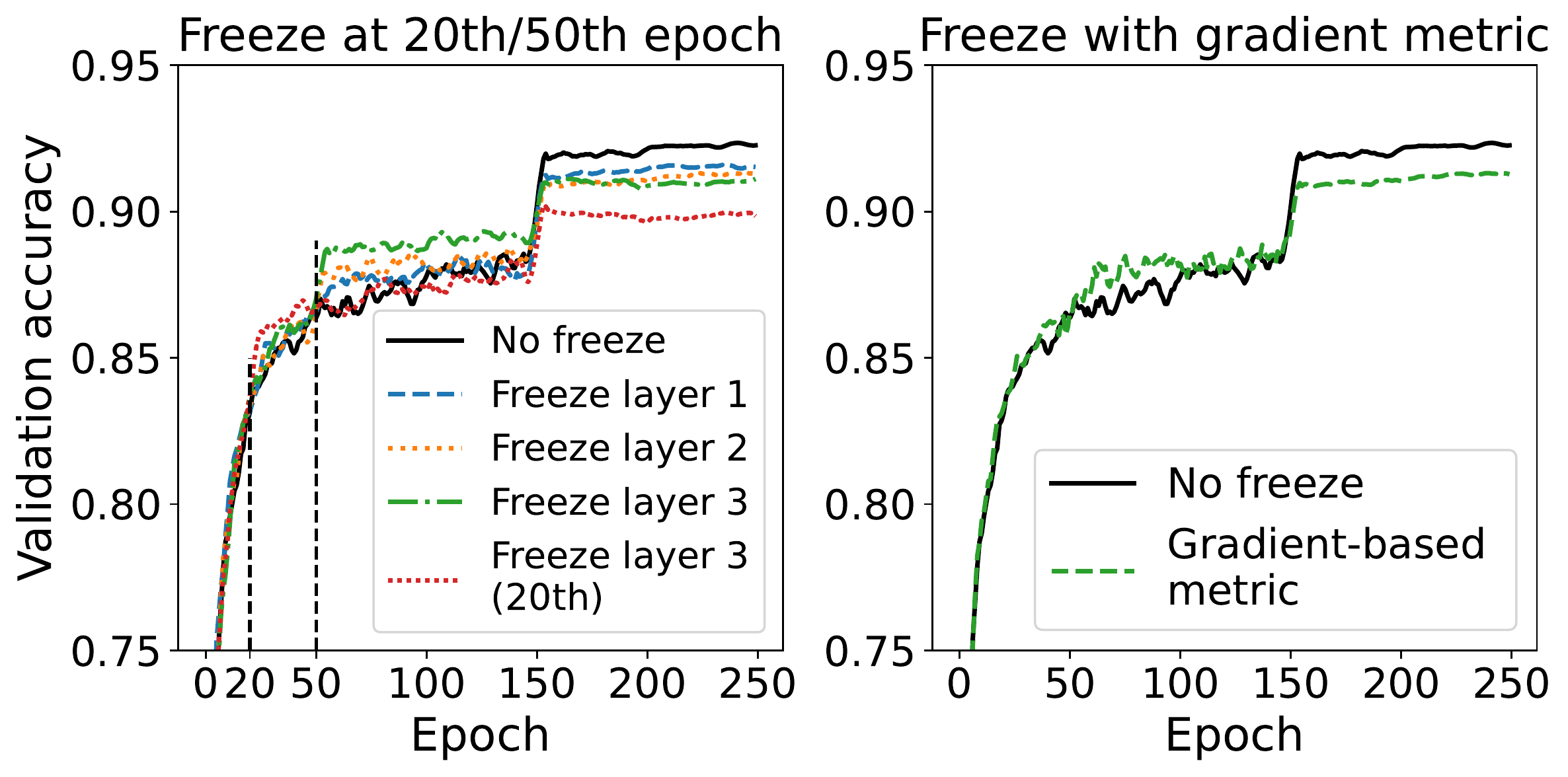}
  \caption{Prematurely freezing layers with transfer learning techniques can hurt the final accuracy in general training.}
  \label{fig:skip_blocks}
\end{figure}

\paragraph{Challenges in layer freezing.} 
Although this opportunity has been pointed out by ML community, it is not feasible to run the hypothetical post hoc analysis (e.g., PWCCA) in practice. 
Quantifying the training progress of a layer is difficult due to the lack of prior knowledge (e.g., a trained model). 
Furthermore, we find that prematurely freezing layers using transfer learning techniques can substantially hurt a model's accuracy. 
To demonstrate this, we investigate the impact of static freezing~\cite{lee2019would} and a gradient-based metric~\cite{liu2021autofreeze}, both from fine-tuning pre-trained models, on the final accuracy when training ResNet-56. 
In Figure~\ref{fig:skip_blocks}, we first fix the parameters of each layer module at the 20th/50th epoch and show their validation accuracies alongside the baseline.
The degraded accuracies indicate that freezing layers prematurely can hurt accuracy by nearly 2\% which is huge for such models.
We test freezing with a gradient-based metric~\cite{liu2021autofreeze} to reach a similar 20\% speedup, but constantly find $\sim$1\% non-negligible accuracy loss.
According to the benchmark~\cite{cifar10leader}, proposing a DNN architecture or training method usually improves the accuracy by less than one percent, so the such loss could offset the benefit of a new ML technique.\footnote{Since AutoFreeze's implementation is deeply coupled with Transformers, we optimize the performance on ResNet training to the best of our ability.}
This motivates an accurate way to freeze layers in general training beyond fine-tuning.
We further compare \sys with existing freezing proposals in \S\ref{sec:related}.
\looseness=-1

\begin{figure*}[t!]
  \centering
  \includegraphics[width=\textwidth]{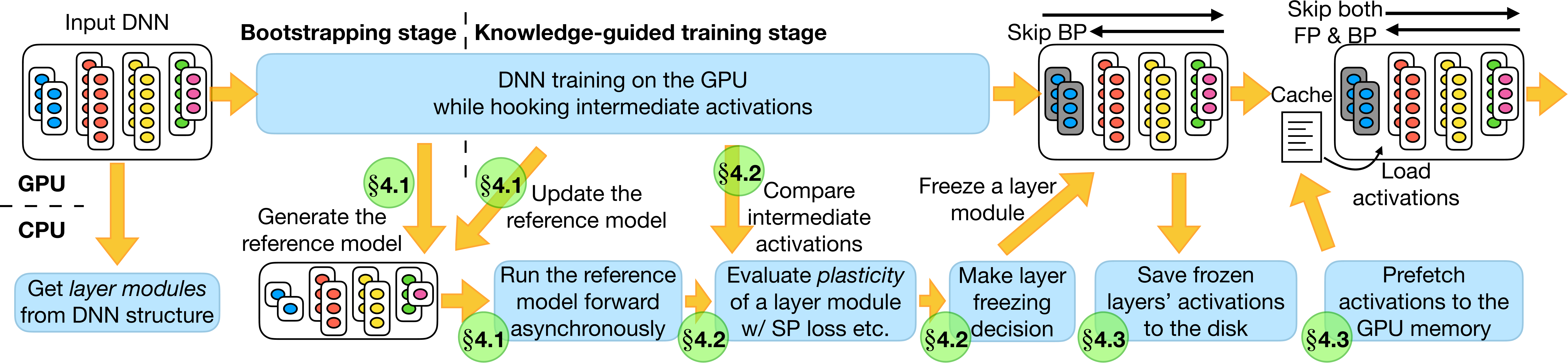}
  \caption{\name overview. \name covers two training stages (bootstrapping and knowledge-guided stages) and has three major design components (generating and executing the reference model, layer freezing with plasticity evaluation, and skipping FP with prefetching).}
  \label{fig:overview}
\end{figure*}
\section{\name Overview}
\label{sec:overview}

We propose \name, an efficient DNN training system that detects and freezes the converged layers in a practical manner. 
Lacking prior knowledge of the hypothetical fully-trained model, \sys introduces a self-generated {\em reference model} during training to provide semantic knowledge for evaluating a layer's convergence with minimal system overhead ($\S$\ref{subsec:ref_model}).
The reference model is essentially an accompanying lightweight DNN with the same architecture as the model being trained to match their internal layers and understand layer-wise performance.

To quantify the training progress, \sys defines a system metric of \emph{plasticity}.
A layer's plasticity is formulated as the difference between the intermediate activation tensors of the training model and its reference model given the same mini-batch input.\footnote{We measure the difference using the Similarity-Preserving loss (SP loss)~\cite{Tung_2019_ICCV}, a novel loss function recently developed to compare two activation tensors for CV tasks and also echoed in NLP~\cite{park2021emnlp}.
Unlike PWCCA~\cite{morcos2018insights} for post hoc convergence analysis, SP loss focuses on capturing the semantic difference for DNN training, making it a perfect fit for plasticity evaluation.}
The plasticity changes as the model evolves over training, and \sys considers the layers with stable plasticity values to be converged, whereby \sys freezes these layers without hurting the accuracy (\S\ref{subsec:measure_maturity}).

\newlength{\twosubht}
\newsavebox{\twosubbox}
\begin{figure}[!t]
  \sbox\twosubbox{%
    \resizebox{\dimexpr\columnwidth-1em}{!}{%
      \includegraphics[height=4cm]{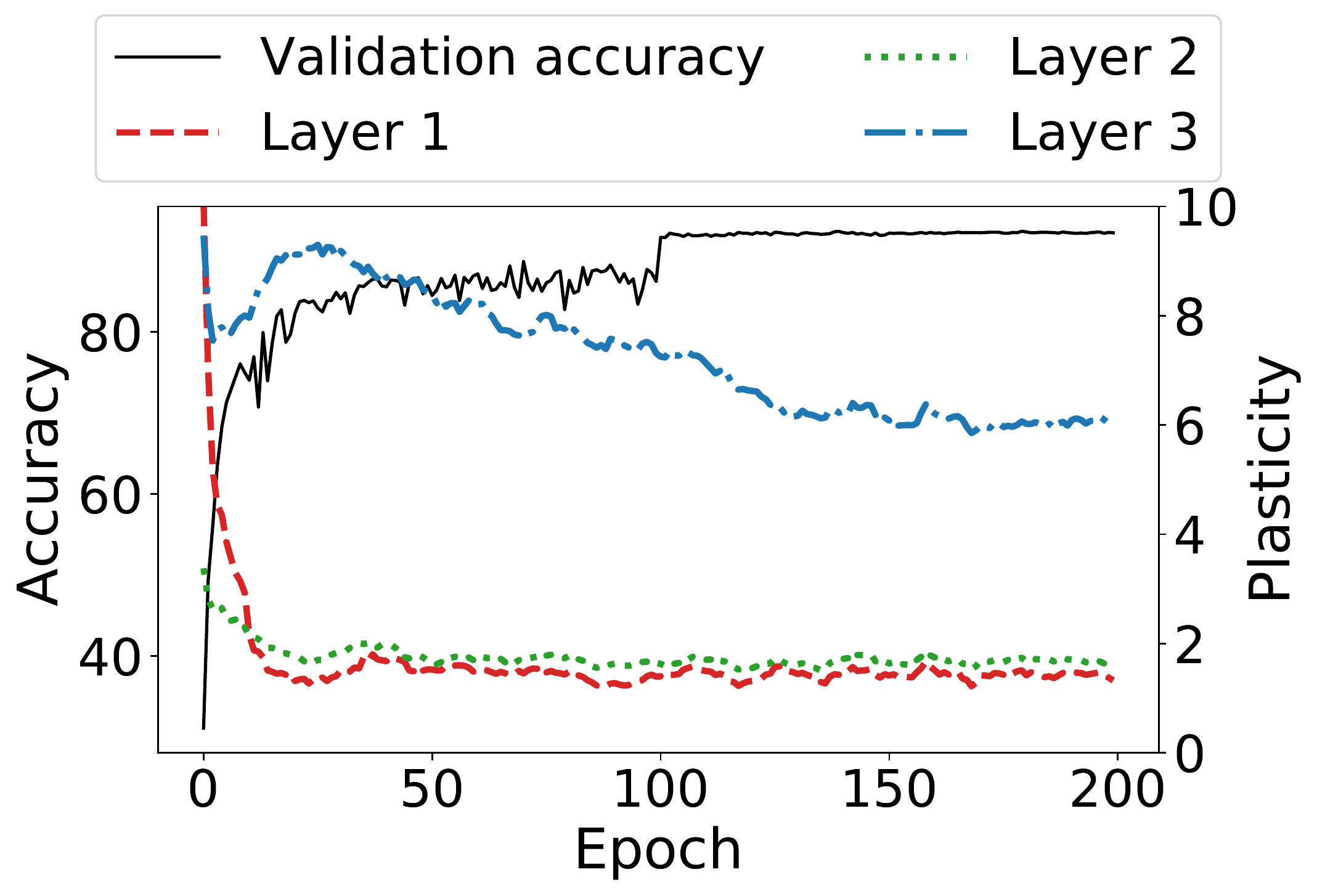}
      \includegraphics[height=4cm]{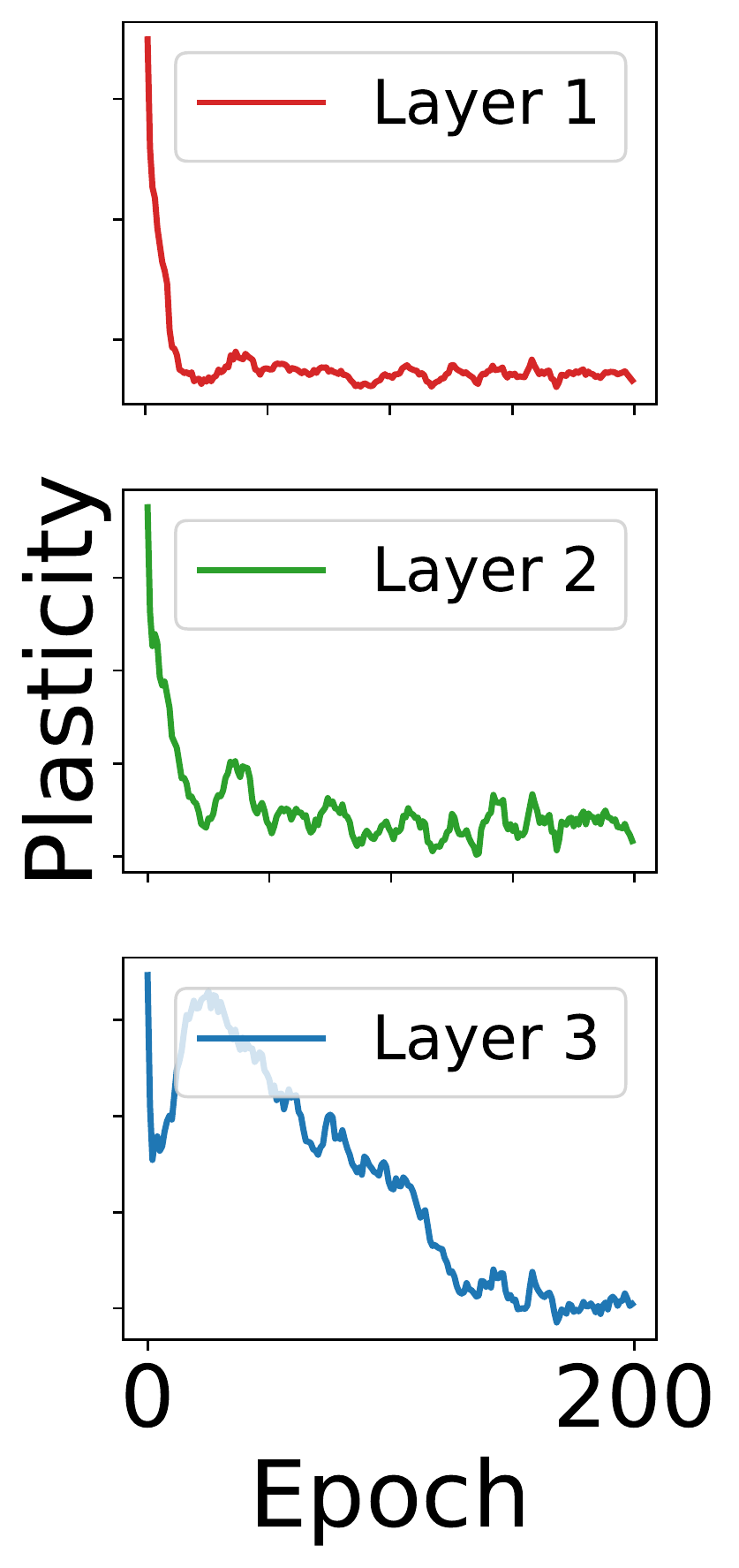}
    }%
  }
  \setlength{\twosubht}{\ht\twosubbox}  
  \centering
  \subcaptionbox{Overall plasticity and accuracy
    \label{fig:cifar10}}{%
    \includegraphics[height=\twosubht]{figures/sim_loss.pdf}
  }\quad
  \subcaptionbox{Trends
    \label{fig:loss_shapes}}{%
    \includegraphics[height=\twosubht]{figures/loss_shapes.pdf}
  }
  \caption{The plasticity of the front layers drops quickly, and they will produce semantically similar activations for most training iterations.}
\end{figure}

To validate the effectiveness of plasticity in capturing the training progress, we use ResNet-56 and generate a reference model with the same architecture but pre-trained for only 50 epochs. We measure the plasticity of ResNet-56's first three layer modules during training. In Figure~\ref{fig:cifar10}, the top black curve shows the validation accuracy, 
and the other three indicate the derived plasticity for each layer module.
We find that, in the first $\sim$30 epochs, the plasticity of the first two modules converges quickly to a low level while the plasticity of layer module 3 is much higher and unstable. These layers show different trends of plasticity, more clearly after normalization in Figure~\ref{fig:loss_shapes}.
We find similar freezable regions when using the post hoc analysis (Figure~\ref{fig:pwcca}), e.g., layer 1 converges near the 50th epoch, while layer 3 only converges at the last epochs.

Plasticity requires no prior knowledge as PWCCA, only a training-in-progress model for reference, and accurately captures the trend of layer convergence.
We conduct correctness analysis and find that plasticity shows similar patterns of layers' convergence as PWCCA but has $\sim$10$\times$ lower overhead and optimized as a loss rather than for visualization (details deferred to the full version). 
The algorithm behind plasticity is tested performant for various tasks compared to traditional gradients and other activation-based metrics~\cite{Tung_2019_ICCV,park2021emnlp}.

\paragraph{Training life cycle with \name.}
Figure~\ref{fig:overview} describes the high-level workflow of \name in two stages.

(1) \emph{Bootstrapping stage}: 
When a job is submitted, \name starts to monitor the job.
The bootstrapping stage is a \emph{critical period} of training, during which the DNN is sensitive and no parameter is eligible for freezing, according to recent research~\cite{achille2018critical}.
\name monitors the changing rate of the training loss (in line with the later plasticity monitoring) and moves to the next stage as the DNN moves out of the critical period.

(2) \emph{Knowledge-guided training stage}: 
\name generates the reference model on the CPU using the latest snapshot of the training model.
Afterwards, \name collects the intermediate activation of the frontmost non-frozen layer of the full model and its reference model for plasticity evaluation (\S\ref{subsec:ref_model}), freezes the layer once it reaches the convergence criteria (\S\ref{subsec:measure_maturity}), and moves to the next active layer.
\name excludes the frozen layers during BP (and parameter synchronization in case of distributed training) to accelerate training.
Meanwhile, \name caches the frozen layer's activations to the disk, so that we can also skip the FP computation by prefetching the intermediate results for the same input (\S\ref{subsec:freezing}).
\section{Design}
\label{sec:design}

Next we dive into the details on how to capture the semantic knowledge during training (\S\ref{subsec:ref_model}), 
with which \name optimizes the computation and communication in the backward pass (\S\ref{subsec:measure_maturity}), as well as the forward pass (\S\ref{subsec:freezing}) on the fly.

\subsection{\name Architecture}
\label{subsec:ref_model}

Directly running another full DNN to measure the internal layers' plasticity can greatly slow down the training.
Instead, \name decouples the control logic and the training logic with a controller-worker abstraction (\S\ref{subsubsec:controller}), and asynchronously performs plasticity evaluation (\S\ref{subsubsec:asyn}).
Besides, \name generates and continuously updates the reference model by fast quantization (\S\ref{subsubsec:quantization}).

\subsubsection{Controller-Worker Framework}
\label{subsubsec:controller}

\begin{figure}[!t]
    \centering
    \includegraphics[width=\columnwidth]{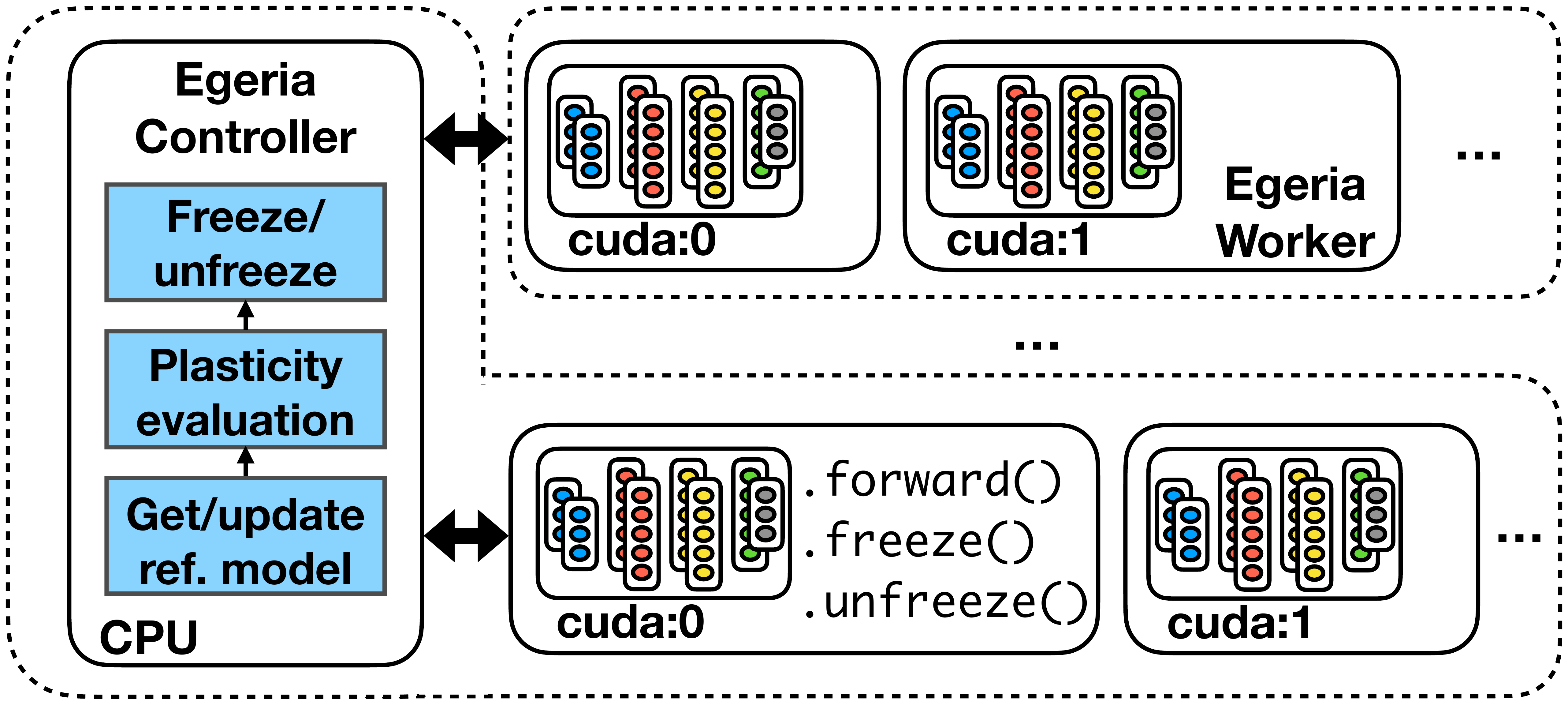}
    \caption{\name uses a controller-worker framework. The controller co-locates on a training node. Solid lines denote device and logical boundaries; dashed lines denote machines.}
    \label{fig:controller}
\end{figure}

Figure~\ref{fig:controller} illustrates the controller-worker framework of \name, 
which primarily consists of a logically centralized controller and workers:
\begin{itemize}
    \item \emph{Controller}:
    It manages the life cycle of the reference model, including its generation and execution, gathering data for plasticity evaluation, and making layer freezing/unfreezing decisions for workers. It makes plasticity evaluations at one place to reduce the overall computation overhead in case of distributed training, as multiple controllers only change the sample size.

    \item \emph{Worker}:
    Each training worker has an \name worker process.
    In addition to the original training operations, it performs \name tasks, including transmitting data and handling controller decisions.
    The updated \texttt{forward()} method uses hooks to obtain the intermediate activation tensors.
    The \texttt{freeze()} and \texttt{unfreeze()} methods will be called by the controller and apply on target layers.

\end{itemize}

\subsubsection{Non-Blocking Plasticity Evaluation}
\label{subsubsec:asyn}

\begin{figure}[!t]
    \centering
    \includegraphics[width=\columnwidth]{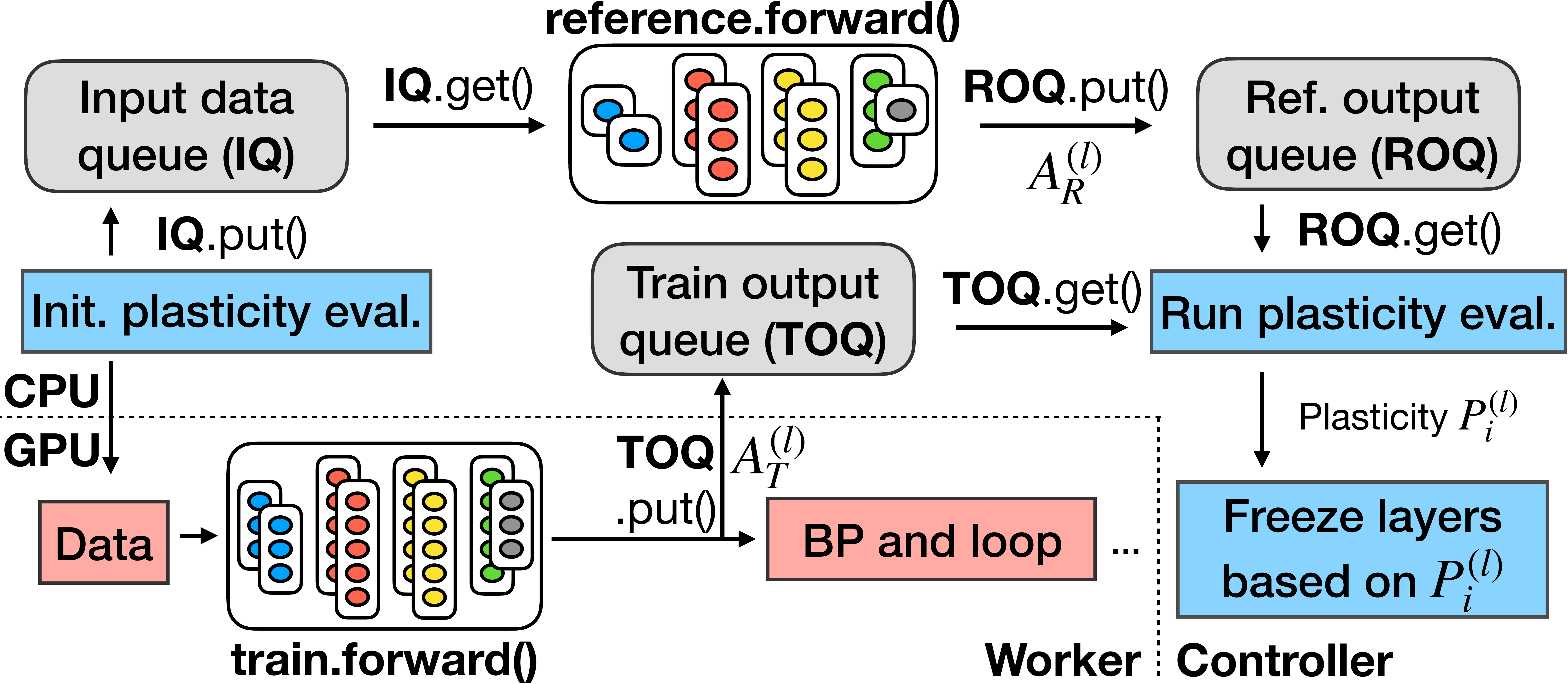}
    \caption{Red and blue blocks are worker and controller operations, respectively. They interact asynchronously through multiprocessing queues (gray blocks).
    }
    \label{fig:queue}
\end{figure}

To avoid slowing down the training, the controller runs an efficient reference model on CPUs in a \emph{non-blocking and asynchronous} fashion, as shown in Figure~\ref{fig:queue}.
ML training servers typically have a high CPU-to-GPU ratio (e.g., 6:1~\cite{awsp4}) since GPUs are the scarcest resource.
In addition, ML system research suggests that GPUs can be fully utilized by exploiting abundant resources like CPUs~\cite{jeon2019analysis,jiang2020unified}, storage~\cite{mohan2020analyzing}, and networking~\cite{peng2019generic}.
Similarly, \sys trades limited CPU cycles which are optimized for int8 inference-only operation~\cite{onednn} for reduced GPU workload.
Since the plasticity evaluation runs periodically (e.g., every $\sim$10 to $\sim$100 iterations) and asynchronously, this non-time-critical process will not interfere with other CPU operations and can be well hidden behind GPU computation, as tested in our evaluation.

We implement the asynchronous plasticity evaluation using three \textit{single-producer/single-consumer queues}: 
(1) When \name initiates a plasticity evaluation, the co-located \name worker puts the data batch in the input queue (IQ).
(2a) The controller process polls IQ, runs a forward pass on the reference model, and puts the hooked intermediate activation $A_{R}^{(l)}$ to the reference output queue (ROQ).
The controller only executes the forward pass at low CPU load (e.g., <50\%) to avoid interference to other CPU-based auxiliary operations, thanks to the non-blocking framework.
(2b) The co-located \name worker puts the hooked intermediate activation $A_{T}^{(l)}$ to the training output queue (TOQ) and continues the training loop without blocking.
(3) The controller polls ROQ and TOQ and calculates the plasticity of the frontmost active layer modules to make freezing decisions (\S\ref{subsec:measure_maturity}).

\subsubsection{Generating and Updating the Reference Model}
\label{subsubsec:quantization}

An ideal reference model should execute fast for diverse models and provide semantically meaningful activations.
To this end, \name generates the reference model for the full model being trained on the fly.

There are several techniques to generate a light-weight version for a large model.
For example, neural architecture search (NAS)~\cite{zoph2016neural} and knowledge distillation (KD)~\cite{hinton2015distilling} can compress the model but have prohibitively large computation overhead.
Besides, they may produce different architectures that do not match the internal layers for plasticity evaluation.
Therefore, \name adopts \emph{post-training quantization}~\cite{han2015deep} to instantly generate a reference model with the same structure.

Quantization is a popular model compression technique to accelerate inference on CPUs~\cite{wu2019machine,hu2020starfish,xie2019source}.
It reduces the precision of model's parameters (e.g., from 32-bit floating-point to 8-bit integers).
By default, \name quantizes the reference model using 8-bit integers.
This can reduce the reference memory footprint by 3$\times$ to 4$\times$ and accelerate the forward pass by 2$\times$ on CPUs; meanwhile, a lower precision (e.g., int4 or int2) cannot further improve the performance due to the CPU instruction set~\cite{lin2020mcunet}.
In our test, int8 quantization reaches a sweet spot to achieve both fast and accurate semantic reference (\S\ref{subsec:sensitivity}).
\sys can fall back to full-precision reference if the training DNN is extremely sensitive or running the reference on GPU in case the CPU resources are scarce, adapting to various environments.
In addition, we design the freezing workflow (Algorithm~\ref{alg:freeze}) with robustness kept in mind.

\name will periodically update the reference model every $W$ iterations (\S\ref{subsubsec:howtounfreeze} discusses the frequency value) using the latest training snapshot to keep it up-to-date for plasticity evaluation.
We empirically find that a stale reference model can amplify the inherent fluctuations in stochastic gradient descent (SGD) training~\cite{bottou2010large}, making the plasticity curve drastically changes, which is hard for \name to understand its trend.
Updating the reference model can smooth the plasticity and better keep up with the baseline.
We use a periodic update because we find that the frequency of reference update is quite insensitive:
Frequently updating is unnecessary because learning is a slow process, while a changing rate-based method that slows down updating in the later training stage brings little gain since the overhead is low anyway.

\subsection{Freezing Layers with Plasticity}
\label{subsec:measure_maturity}

Next, we elaborate how \name decides when to freeze the stable layers by comparing the intermediate activations of the model being trained with that of the reference model.
During this stage, \name needs to address two challenges:
\begin{enumerate*}[(1)]
    \item how to quantify the training plasticity of a layer module; and
    \item how to accurately make the layer freezing decision.
\end{enumerate*} 

\subsubsection{Evaluating the Plasticity of a Layer Module}
\label{subsubsec:plasticity}
The raw data we collect is the intermediate activation tensors of the training and reference models, which have been widely studied as a direct metric of layer performance in understanding and exploring new methods of DNN training~\cite{morcos2018insights,NIPS2014_5347}.
For example, knowledge distillation uses the difference of activations between the training model and a trained teacher model~\cite{Tung_2019_ICCV,romero2014fitnets} as a supervisory signal to improve accuracy because intermediate activation plays an important role in forming the decision boundaries for the partitioning of the feature space in each hidden layer~\cite{pan2016expressiveness,heo2019knowledge}.
Rather than comparing the gradients of the two models calculated against a hard label, using intermediate activation as a ``soft'' label is proven more effective and accurate in guiding training in recent research~\cite{aguilar2020knowledge,Tung_2019_ICCV} since it is a more semantically meaningful indicator and provides \emph{contextual} knowledge~\cite{coenen2019visualizing,park2021emnlp}.
We also test freezing with a gradient-based metric~\cite{liu2021autofreeze} from fine-tuning but find $\sim$1\% accuracy loss, which is undesirable for a general training system and echoes the KD research, as shown in Figure~\ref{fig:skip_blocks}.
To freeze layers accurately, we quantify the training plasticity by measuring the changes in the layer's intermediate activations.

We use the SP loss~\cite{Tung_2019_ICCV} that shows high efficacy in KD tasks to compare the intermediate tensors between the two models.
The theory behind SP loss is that the same input data will elicit similar pair-wise activation patterns in trained models for both CV and NLP tasks~\cite{Tung_2019_ICCV,park2021emnlp}.
Different from gradient norm that is calculated against single-dimension hard labels (e.g., ``cat'' for image classification), SP loss calculated from two high-dimension activation tensors can better capture the semantic and contextual information, as discussed in recent KD research~\cite{aguilar2020knowledge,Tung_2019_ICCV} and tested in \S\ref{subsec:end2end}.
Thus, \sys can make freezing decisions more accurately, and our evaluation shows higher accuracy when achieving the same speedup.
Compared to PWCCA~\cite{morcos2018insights} used in post hoc analysis, our empirical analysis finds that SP loss shows similar training progress of layers and freezing opportunities.
We choose SP loss because
\begin{enumerate*}[(1)]
    \item it is designed as a training loss to measure the performance difference \emph{in the actual scale} for direct model updating, while PWCCA is a visualization and analysis tool that uses \emph{weighted} projection to fit largely distinct values into the scale of 0\textendash1, showing performance gaps differently in different training stages;
    \item we find that PWCCA is more computation-intensive for its projection operation.
\end{enumerate*}

We focus on the performance of a layer module that contains consecutive layers defined together.
Layers in a module are closely related to performing a sequence of transformations for a certain goal~\cite{Li_2020_CVPR} and have similar training progress.
Meanwhile, individual layers with fewer parameters (e.g., linear layers) are less stable in SGD training.
Though it's rare, even a few individual front layers might not converge in strict order (as observed in \cite{zhang2019all}), our module-based freezing can mitigate this and revisit them in the future unfreezing stage (\S\ref{subsubsec:howtounfreeze}).
\name provides configuration options to customize the granularity of layer module through regular expression, e.g., evaluating every convolutional layer.

Given the input data of batch size $b$, we denote the activation tensors of the training and reference models at a layer $l$ as $A_{T}^{(l)}$ and $A_{R}^{(l)}$.
For the image data, the activation tensors $A_{T}^{(l)},A_{R}^{(l)}{\in}{\mathbb{R}^{b{\times}c{\times}h{\times}w}}$, where $c$, $h$, and $w$ are channel number, height, and width; similar for the word embeddings.
Then SP loss will align $A_{T}^{(l)}$ and $A_{R}^{(l)}$ to $b{\times}b$-shaped matrices, which encode the pair-wise similarity in the activation tensors that are elicited by the input mini-batch.
We then denote the \emph{training plasticity} $P_{i}^{(l)}$ of layer $l$ at an iteration $i$ using the SP loss between the two matrices, representing the semantic difference compared to the reference model, as shown in Equation~\ref{eq:plasticity}.
The lower and more stable the plasticity, the DNN layers are closer to convergence.

\begin{equation}
    \label{eq:plasticity}
    P_{i}^{(l)} = {SP\_loss}(A_{T}^{(l)}, A_{R}^{(l)})
\end{equation}

\subsubsection{How to Decide Layer Freezing}
\label{subsubsec:howtounfreeze}
During the knowledge-guided training stage, \name will periodically run the plasticity evaluation every $n$ iterations and decide whether to freeze the layer or not.
The intuition of the freezing criterion is straightforward: If a layer's plasticity becomes stationary for some iterations $W$, \name considers its semantic performance stable and can safely freeze it.

When obtaining the plasticity $P_{i}^{(l)}$, we first smooth it with the moving average of its recent values (using $W$ or the max span as the history buffer size), as shown in Equation~\ref{eq:average}.

\begin{equation} \label{eq:average}
    \overline{P_{i}^{(l)}} = 
    \begin{cases}         
        \frac{{{P}_{i-W}^{(l)}+\cdots+{{P}_{i}^{(l)}}}}{W}, & i\geq{W} \\
        \frac{{{P}_{0}^{(l)}+\cdots+{{P}_{i}^{(l)}}}}{i}, & i<{W}
    \end{cases}
\end{equation}

To determine whether the curve has become stable, we fit $\overline{P_{i}^{(l)}}$ with linear least-squares regression to a straight line and analyze its slope (0 means no change at all).
This method can filter out the drastic fluctuation in SGD training and provide a recent history context than simply evaluating the delta.
If the plasticity slope has been less than the tolerance $T$ for $W$ evaluations, we consider the layer converged, freeze it, and move to the next layer, as detailed in Algorithm~\ref{alg:freeze}.
This simple yet effective method has a similar intuition to the early stopping in DNN training~\cite{es_keras, es_scikit}.

\name monitors the frontmost active layer module $l$ to avoid a fragmented frozen model.
According to the chain rule of automatic differentiation~\cite{maclaurin2015autograd}, only excluding the last link of backpropagation can reduce the workload.
It is widely recognized that the front layers converge faster~\cite{NIPS2014_5347,zeiler2014visualizing,shen2020reservoir,rogers2020primer}, and \name can handle exceptions with the aforementioned module-based freezing and unfreezing mechanism.

\paragraph{Unfreezing.}

Learning rate (LR) scheduling can influence the convergence of all layers~\cite{achille2018critical} (e.g., the PWCCA score and accuracy boost in Figure~\ref{fig:pwcca} and \ref{fig:skip_blocks}) and is an external factor to the model.
LR annealing~\cite{cs231n_lr}, the most commonly employed scheduling technique, recommends gradually lowering the LR during training with step decay or exponential decay.
Given this, \name will restart training all the frozen layers if the LR has dropped over a factor of 10 since the frontmost layers' freeze and halve the counter and history buffer $W$ for refreezing, which we find effective for different models.
Another type of LR scheduling is periodically increasing and decreasing the LR, e.g., cosine annealing~\cite{loshchilov2016sgdr} and cyclical LR~\cite{smith2017cyclical}.
Due to its complexity, \name lets the user customize the unfreezing and refreezing criteria, e.g., training for a few iterations in each cycle after freezing a layer.

\paragraph{Hyperparameters guideline.}
We use three hyperparameters: $n$ (plasticity evaluation and bootstrapping stage monitoring interval), $T$ (the tolerance of plasticity slope), and $W$ (number of low slope evaluations to freeze layers and history buffer).
They are highly related and can be automatically set with some task knowledge.
\sys sets $T$ for each layer module as the 20\% of the maximal plasticity slope in its initial 3 readings; the rationale is that layers move differently and thus should have per-layer thresholds.
We recommend setting $n$ as a moderate frequency value that can cover the evaluation of all layers.
For example, for training ResNet-56 with 7 layer modules, LR scheduling, and $W$=10 for 200 epochs ($\sim$78k iterations), we set $n$ to 300 iterations ($\approx78k/(10*2)/7/(1+0.5+0.25)$ considering bootstrapping, smoothing delay, and window halving).
Our extensive empirical analysis shows that we achieve consistently good performance across different parameters following our general guideline, and we conduct sensitivity analysis of $W$, $n$, and $T$ in \S\ref{subsec:sensitivity} to show their impact on performance with largely different values.
The changing rate of ending the bootstrapping stage is permissively set to 10\%.

\begin{algorithm}[!t]
    \SetKwFunction{isOddNumber}{isOddNumber}
    \SetKwInput{Input}{Input}
    \SetKwInput{Output}{Output}
    \SetKw{And}{and}

    \SetKwFunction{freezeLayer}{freezeLayer}
    \SetKwFunction{unfreezeAllLayers}{unfreezeAllLayers}
    \SetKwFunction{customizedUnfreeze}{customizedUnfreeze}
    \SetKwFunction{checkPlasticity}{checkPlasticity}
    \SetKwFunction{calculateSPLoss}{calculateSPLoss}
    \SetKwFunction{smoothPlasticity}{smoothPlasticity}
    \SetKwFunction{linearRegression}{linearRegression}
    \SetKwFunction{windowLinearFit}{windowLinearFit}
    \SetKwProg{Fn}{Function}{:}{}

    \newcommand\mycommfont[1]{\footnotesize\textcolor{blue}{#1}}
    \SetCommentSty{mycommfont}

    \SetKwComment{Comment}{$\triangleright$\ }{}

    \KwIn{

        Intermediate activations of the training and reference models $A_{T}^{(l)}$, $A_{R}^{(l)}$,
        layer module $l$, training iteration $i$, counter and history buffer length $W$, tolerance $T$.
    }

    \KwOut{
        The (updated) frontmost active layer $l$.
    }
    \tcc{Initialize global variables.}

    $pList_{l}$ $\leftarrow$ $\emptyset$ \Comment*[r]{Plasticity evaluation history of $l$ across iterations.}

    $staleCounter$ $\leftarrow$ 0 \Comment*[r]{Number of consecutive stale $\overline{P_{i}^{(l)}}$.}

    \Fn{\checkPlasticity{$A_{T}^{(l)}$, $A_{R}^{(l)}$, $l$, $i$, $T$, $W$}}{

        \KwSty{assert} $l$ is not the last layer

        \eIf{$staleCounter < W$}{

            ${P_{i}^{(l)}}$ $\leftarrow$ \calculateSPLoss($A_{T}^{(l)}$, $A_{R}^{(l)}$, $l$, $i$)

            \tcc{Use moving average to mitigate outliers (Equation~\ref{eq:average}).}

            $\overline{P_{i}^{(l)}}$ $\leftarrow$ \smoothPlasticity{${P_{i}^{(l)}}$, $W$}

            \tcc{Update the time-series plasticity list.}
            $pList_{l}$ $\leftarrow$ $pList_{l}$ $\cup$ $\overline{P_{i}^{(l)}}$
            
            \tcc{Calculate the slope of the linear-fitted plasticity.}
            $s$ $\leftarrow$ \windowLinearFit{$pList_{l}$, $W$}.slope

            \tcc{If the fitting line is close to horizontal.}
            \eIf{$s$ < $T$}{
                $staleCounter \leftarrow staleCounter + 1$
            }{
                $staleCounter$ $\leftarrow$ 0
            }
        }
        {
            \freezeLayer($l$)

            $l \leftarrow$ $l$ + 1
        }
        \tcc{Learning rate-based unfreezing mechanism.}
        \eIf{LR annealing \And LR decreased by 90\%}{
            \unfreezeAllLayers()
            
            $l \leftarrow$ $l_0$ \Comment*[r]{Reset $l$.}
        }
            {\If{cyclical LR scheduling}{
            \customizedUnfreeze()
            }
            }
        \KwRet{$l$}
    }

    \caption{Layer freezing algorithm.}
    \label{alg:freeze}
\end{algorithm}

\subsection{Skipping Forward Pass with Caching and Prefetching}
\label{subsec:freezing}

By freezing the converged front layers, we can exclude them from the backward pass and parameter update to reduce training cost.
However, the forward pass is still necessary because the deep layers require the frozen layers' activations as input~\cite{maclaurin2015autograd}.
Naturally, we can cache the frozen layers' intermediate activations to save the forward pass because they output the same activation given a certain input.

There are two challenges of caching computation results for a DNN training task.
First, training a large model requires a large dataset (e.g., the training set of ImageNet is over 100 GB).
The size of the intermediate activation tensor depends on the output shape of the last frozen layer.
In our evaluation, the storage space of ResNet-50 intermediate activations ranges from 1.5$\times$ to 5.3$\times$ of the input data.
It is not technically appropriate to cache a whole epoch's results to the GPU/CPU memory.
Second, given the memory limit, caching systems usually improve their hit rate by keeping the most frequent content using replacement policies like LRU (least-recently-used).
However, in DNN training, the data loader randomly samples a mini-batch, meaning there is no popular data to prioritize in the cache.

To solve these challenges, we exploit a training workflow feature: Before an iteration, the \emph{data loader} samples future mini-batches in advance, so unlike typical cache systems, we actually ``know the future'' (the incoming data indices)!
Prefetching is an effective technique in ML applications~\cite{bai2020pipeswitch,prefetch_pytorch}.
\name saves the forward computation results of the frozen layers to the disk and prefetches the relevant activations to the GPU memory so that the active deep layers can instantly read them as input.
The cache only stores the recent five mini-batches for minimal memory usage.
Users can set the storage limit for activations that are up to an epoch (see analysis in \S\ref{subsec:overhead}).
At the early training stage, we disable prefetching if the forward pass of a few layers is faster.
In this way, we can skip the forward computation of the frozen layers and efficiently overlap the slower disk access with prefetching.

\begin{figure}[!t]
    \centering
    \includegraphics[width=\columnwidth]{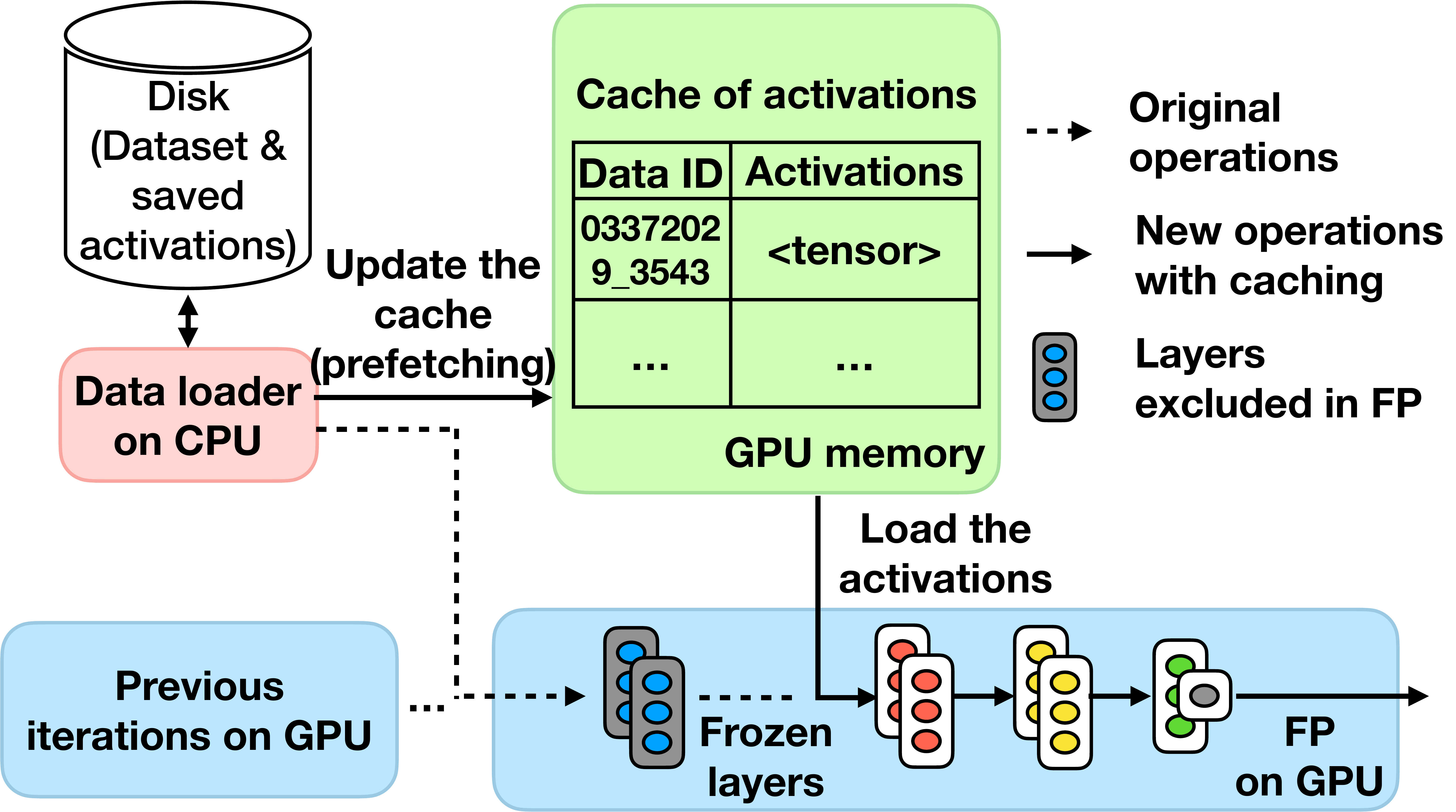}
    \caption{\name caches the intermediate activation and prefetches the tensors into GPU memory during the FP.}
    \label{fig:cache}
\end{figure}

Figure~\ref{fig:cache} illustrates a forward pass in \name.
We maintain a hash table in the GPU memory.
The key is the sample ID, and the value is the corresponding activation tensor.
During training, the data loader will sample a data batch and prefetch their activations from the storage to the GPU memory in parallel to the GPU training.
\name prefetches more than one mini-batch of the future activations, similar to the data loader~\cite{prefetch_pytorch}, depending on the memory and CPU availability.

\name can cover different training techniques and cache their outputs.
\name is compatible with stateless random operations, which are recommended for random data augmentation and dropout~\cite{tf_augmentation,tf_dropout}.
Thus we can deterministically keep the randomly augmented images the same across epochs without ML performance penalty.
Each worker machine maintains its own cache in distributed training to avoid extra network overhead; besides, the random seed is device-dependent.
For most layers, e.g., convolutional (for CNNs) and linear (for language models), the output activation only depends on the parameters given the same data, so caching can work naturally.
For batch normalization layers, the activation also depends on the specific data batch.
\name handles this case using the practice in transfer learning~\cite{fine_tune_keras}: we set these layers to the inference mode, using the dataset statistics to normalize the input rather than the specific batch.
\section{Implementation}
\label{sec:implementation}

\name is independent of DNN training frameworks.
In this paper, we implement and evaluate \name using PyTorch and Huggingface Transformers~\cite{wolf-etal-2020-transformers}.
All the technical dependencies of \name (e.g., quantization and asynchronous computation) can work in other ML frameworks like TensorFlow and MXNet.
\name obtains the layer modules by parsing the model definition and adds forward hooks to obtain intermediate activations.

\paragraph{Reference model.}
When \name controller generates or updates the reference model, it directly moves a snapshot of the training model from GPU to CPU and runs int8 quantization using PyTorch's built-in library.
We use dynamic quantization for NLP models and static quantization for convolutional networks, which add little overhead in the background.
We add the same forward hooks to the reference model to match the training model.

\paragraph{Knowledge-guided training.}
Our high-level API allows us to take advantage of the framework engines to execute all the DNN computation operations.
To freeze a layer, we essentially set the \texttt{requires\_grad} flag of all its parameters to false to exclude the subgraph from gradient computation~\cite{autograd_pytorch}.
Distributed training requires rebuilding the communication buffer.
For caching, we use the dictionary data structure with $O(1)$ lookups.

\begin{table*}[t]
    \centering
    \begin{tabular}{lllllll}
    \toprule
    Task & Model & Dataset & \makecell[l]{Accuracy \\ target} & \makecell[l]{\# Servers $\times$ \\ \# GPUs/server} & \makecell[l]{\# Building \\ layer modules} & \makecell[l]{TTA \\ speedup}  \\
    \midrule
    \multirow{4}*{\makecell[l]{Image \\ classification}} & \multirow{2}*{ResNet-50~\cite{he2016identity}} & \multirow{3}*{ImageNet~\cite{deng2009imagenet}} & \multirow{2}*{\makecell[l]{Top 1\\75.9\%}} & 1$\times$2 & \multirow{2}*{48 (residual blocks)} & 28\% \\
     & & & & 2$\times$2 - 5$\times$2 & & 27\%-33\% \\
     \cmidrule[0.5pt](lr{1em}){2-2}
     & MobileNet V2~\cite{sandler2018mobilenetv2} & & 71.2\% & 1$\times$2 & \makecell[l]{17 (inverted \\ residual blocks)} & 22\% \\ 
     \cmidrule[0.5pt](lr{1em}){2-7}
     & ResNet-56~\cite{he2016identity} & CIFAR-10~\cite{cifar} & 92.1\% & 1$\times$2 & 54 (residual blocks) & 23\% \\
    \midrule
    \makecell[l]{Semantic \\ segmentation} & DeepLabv3~\cite{chen2017rethinking} & VOC~\cite{Everingham10} & \makecell[l]{mIoU \\ 63.3\%} & 1$\times$2 & \makecell[l]{49 (residual blocks \\ and DeepLab head)} & 21\% \\
    \midrule
    \multirow{3}*{\makecell[l]{Machine \\ translation}} & \multirow{2}*{Transformer-Base~\cite{vaswani2017attention}} & \multirow{3}*{\makecell[l]{WMT16\\ EN-DE~\cite{bojar2016findings}}} & \multirow{2}*{\makecell[l]{Perplexity \\ 4.7}} & 4$\times$2 & \multirow{2}*{\makecell[l]{12 (6 encoders \\ \& 6 decoders)}} & 43\% \\
     & & & & 2$\times$2 - 5$\times$2 & & 33\%-43\% \\
     \cmidrule[0.5pt](lr{1em}){2-2}
     & Transformer-Tiny & & 53.3 & 1$\times$8 & 4 (2 \& 2) & 19\% \\
    \midrule
    \makecell[l]{Question \\ answering} & \makecell[l]{BERT-Base~\cite{devlin2018bert} \\ (fine-tuning)} & SQuAD 1.0~\cite{rajpurkar2016squad} & \makecell[l]{F1 score \\ 87.6} & 1$\times$2 & \makecell[l]{12 (Transformer \\ blocks)} & 41\% \\
    \bottomrule
    \end{tabular}
    \caption{Summary of evaluation tasks. Accuracy targets are the converged accuracy in baseline training. \name accelerates different models by 19\%-43\% to reach the target accuracy.}
    \label{table:eval_summary}
\end{table*}

\section{Evaluation}
\label{sec:evaluation}

In this section, we evaluate the effectiveness of \name, namely accelerating DNN training while maintaining accuracy, for different tasks and models using single or multiple machines.
The main takeaways are:
\begin{itemize}
  \item \name can work for different CV and NLP models;
  \item \name accelerates general training by 19\%-43\% without hurting accuracy; and
  \item \name minimizes the system overhead while accurately freezing layers.
\end{itemize}

\subsection{Methodology}

\paragraph{Testbed setup.}
We evaluate \name using two testbed configurations: a cluster of 5 machines and a multi-GPU machine.
In the 5-node cluster, each machine has 2 NVIDIA V100 GPUs (32 GB), 40 CPU cores, 128 GB memory, and 2 Mellanox CX-5 NICs.
To match the CPU-to-GPU ratio of the cloud training instance~\cite{awsp4}, we use 12 CPU cores (equivalent to 24 vCPUs) with \texttt{taskset}~\cite{taskset}.
The testbed has a leaf-spine topology with two core and two top-of-rack (ToR) switches; each ToR switch is connected to 5 servers using 40 Gbps and two core switches using 100 Gbps links.
The single node has 8 NVIDIA RTX 2080 Ti and 64 CPU cores.

\paragraph{Tasks, models, and datasets.}
We evaluate two CV and two NLP tasks: image classification, semantic segmentation, machine translation, and question answering; the corresponding 7 models and 5 datasets are listed in Table~\ref{table:eval_summary}.
We follow the recommended learning rate and batch size settings~\cite{torchvision,ott2019fairseq} and the learning rate schedulers are step decay LR schedule for CV training, inverse square root schedule for Transformer training, and linear schedule for fine-tuning BERT.
We use the all-reduce parameter synchronization scheme for data parallel distributed training with multiple GPUs or machines and allocate one GPU per process.

\paragraph{Metrics and baselines.}
The training performance metric is the time taken to a converged validation accuracy (TTA), as listed in Table~\ref{table:eval_summary}.
We compare \name with the vanilla training framework, PyTorch, and a communication scheduling system, ByteScheduler~\cite{peng2019generic}, in multi-node distributed training using its default configuration.
ByteScheduler achieves the theoretically optimal scheduling without skipping any parameter and full accuracy.
We use scheduling/pipelining systems as the main baseline since maintaining accuracy is our major goal.
We also compare \sys to a recent gradient-based layer freezing system, AutoFreeze~\cite{liu2021autofreeze}, and to using the metric of Skip-Conv~\cite{habibian2021skip} as an alternative to plasticity.
We use the input-norm gate of Skip-Conv, which applies to intermediate activation rather than convolution-specific.

\subsection{End-to-End Training Performance}
\label{subsec:end2end}

\begin{figure*}[!t]
    \begin{subfigure}{0.24\textwidth}
      \includegraphics[width=\linewidth]{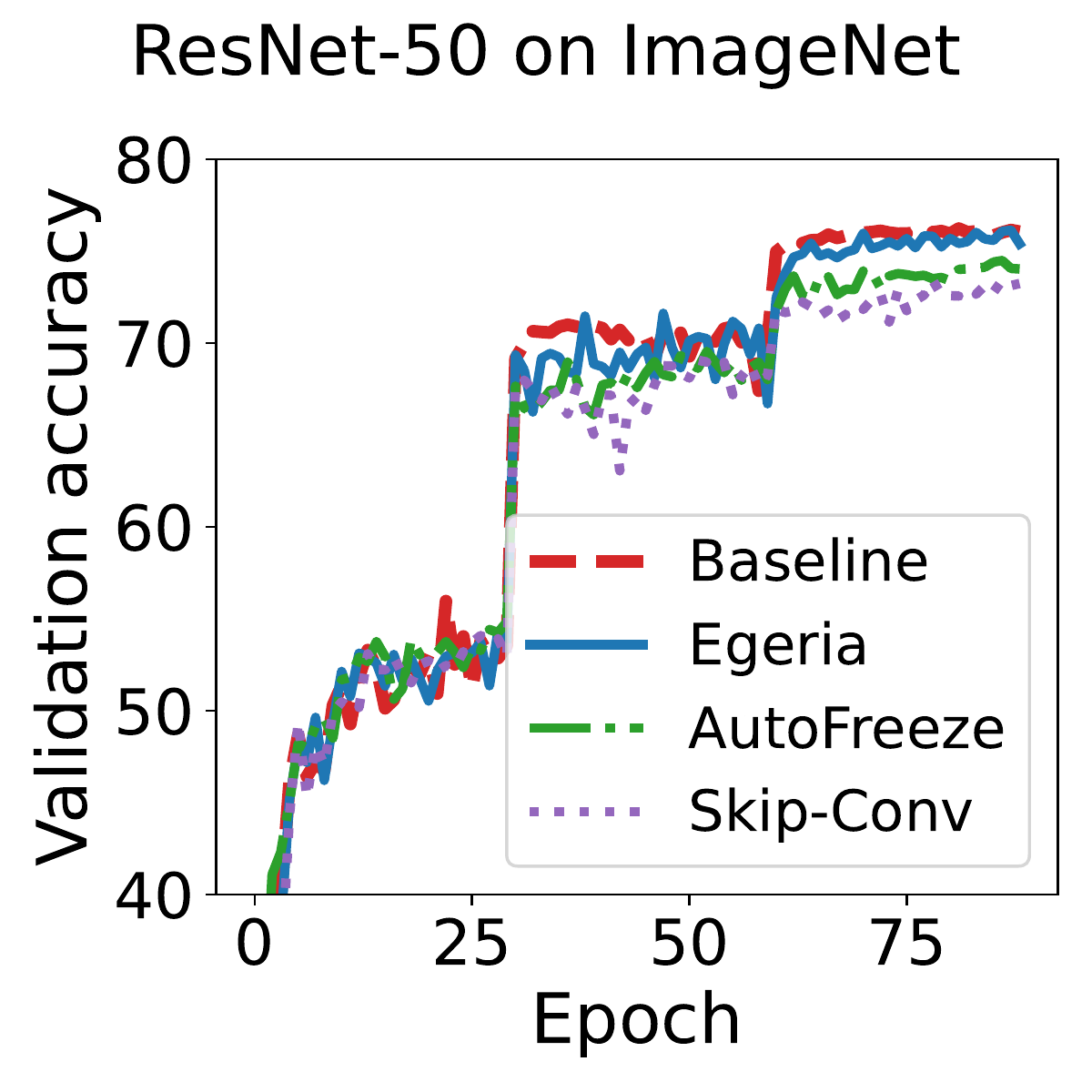}
      \caption{ResNet-50 training achieves 1.5\%+ higher accuracy.}
      \label{fig:resnet50}
    \end{subfigure}\hfill
    \begin{subfigure}{0.24\textwidth}
      \includegraphics[width=\linewidth]{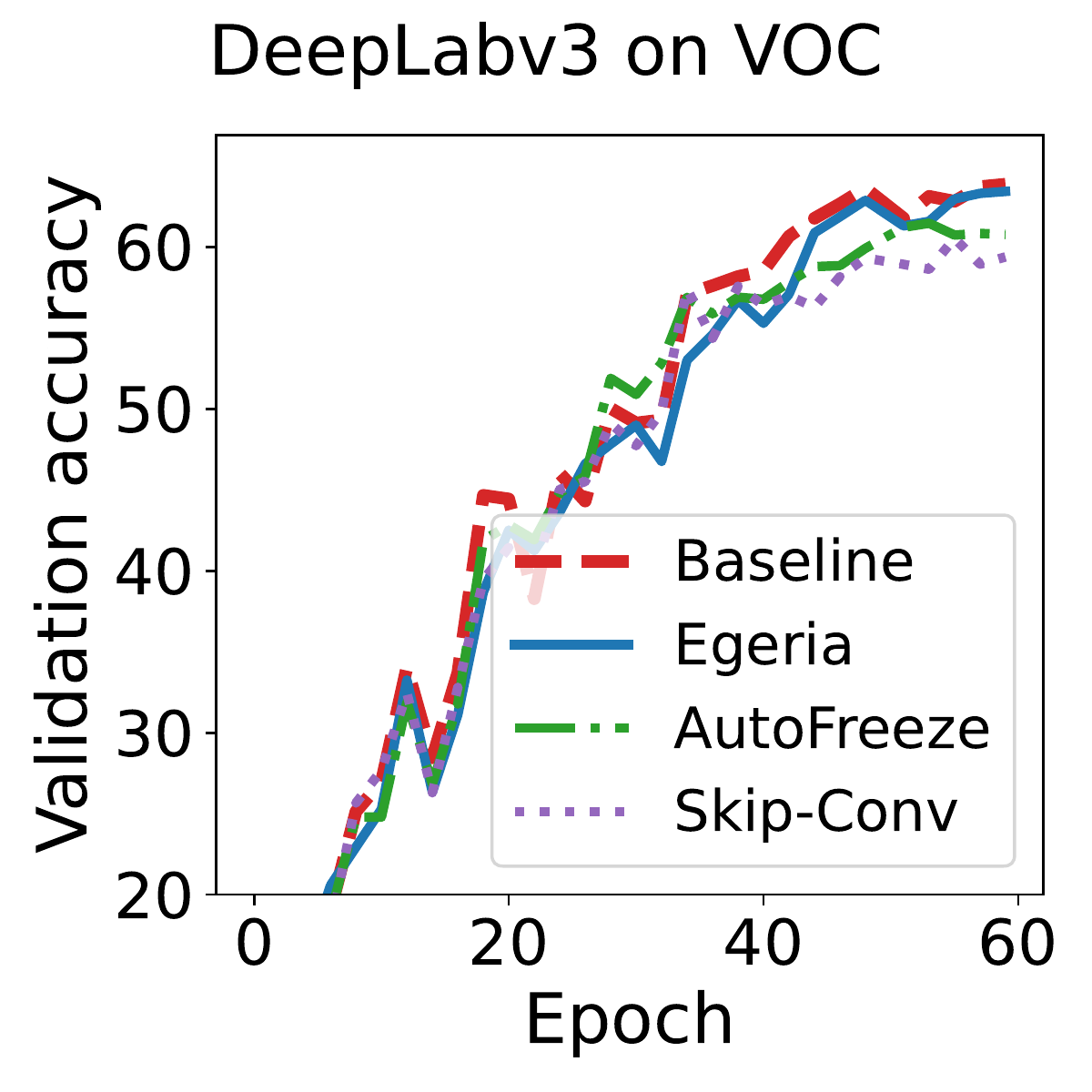}
      \caption{DeepLabv3 training achieves a 21\% speedup and full accuracy.}
      \label{fig:deeplabv3}
    \end{subfigure}\hfill
    \begin{subfigure}{0.24\textwidth}%
      \includegraphics[width=\linewidth]{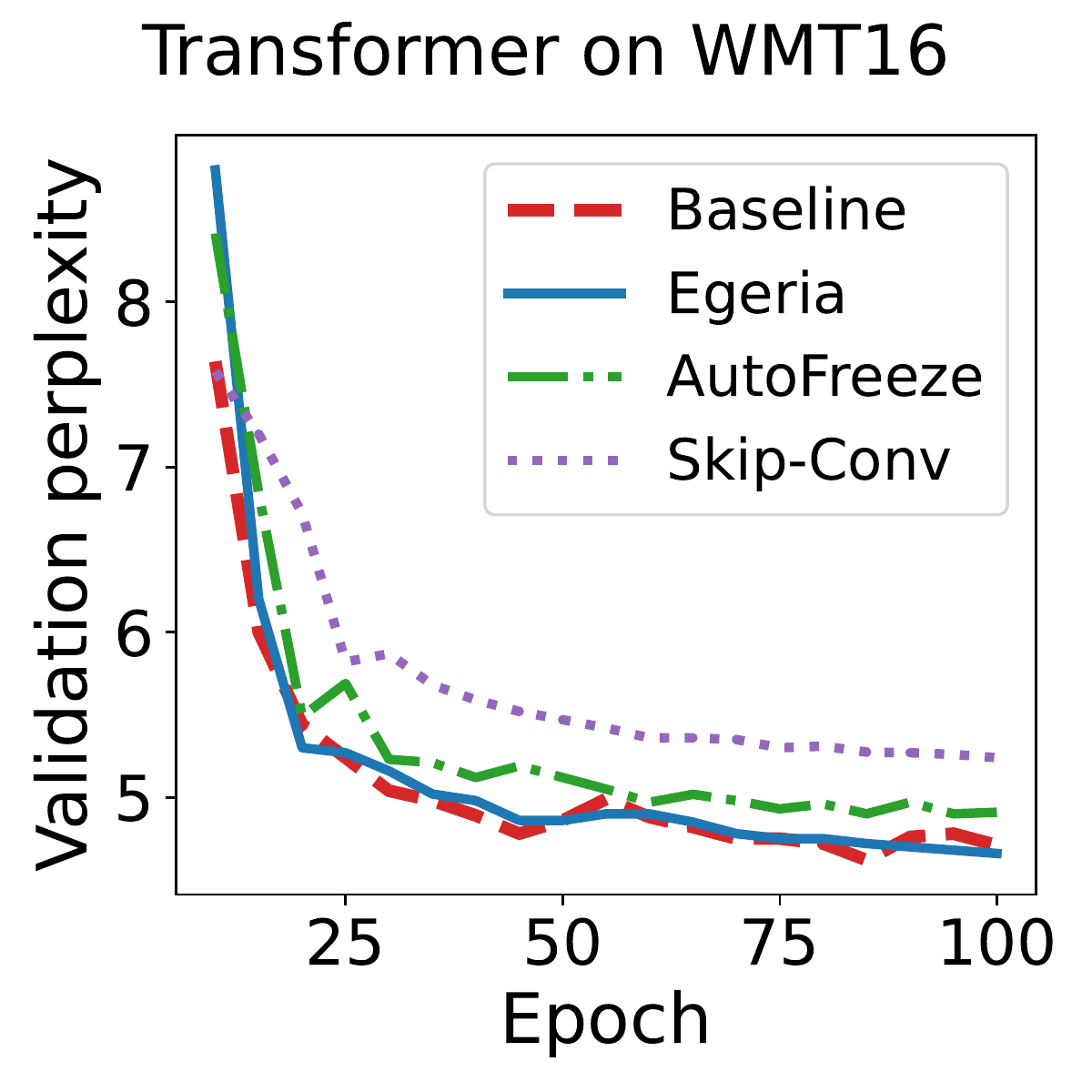}
      \caption{Transformer-Base for machine translation.}
      \label{fig:nmt}
    \end{subfigure}\hfill
    \begin{subfigure}{0.24\textwidth}%
      \includegraphics[width=\linewidth]{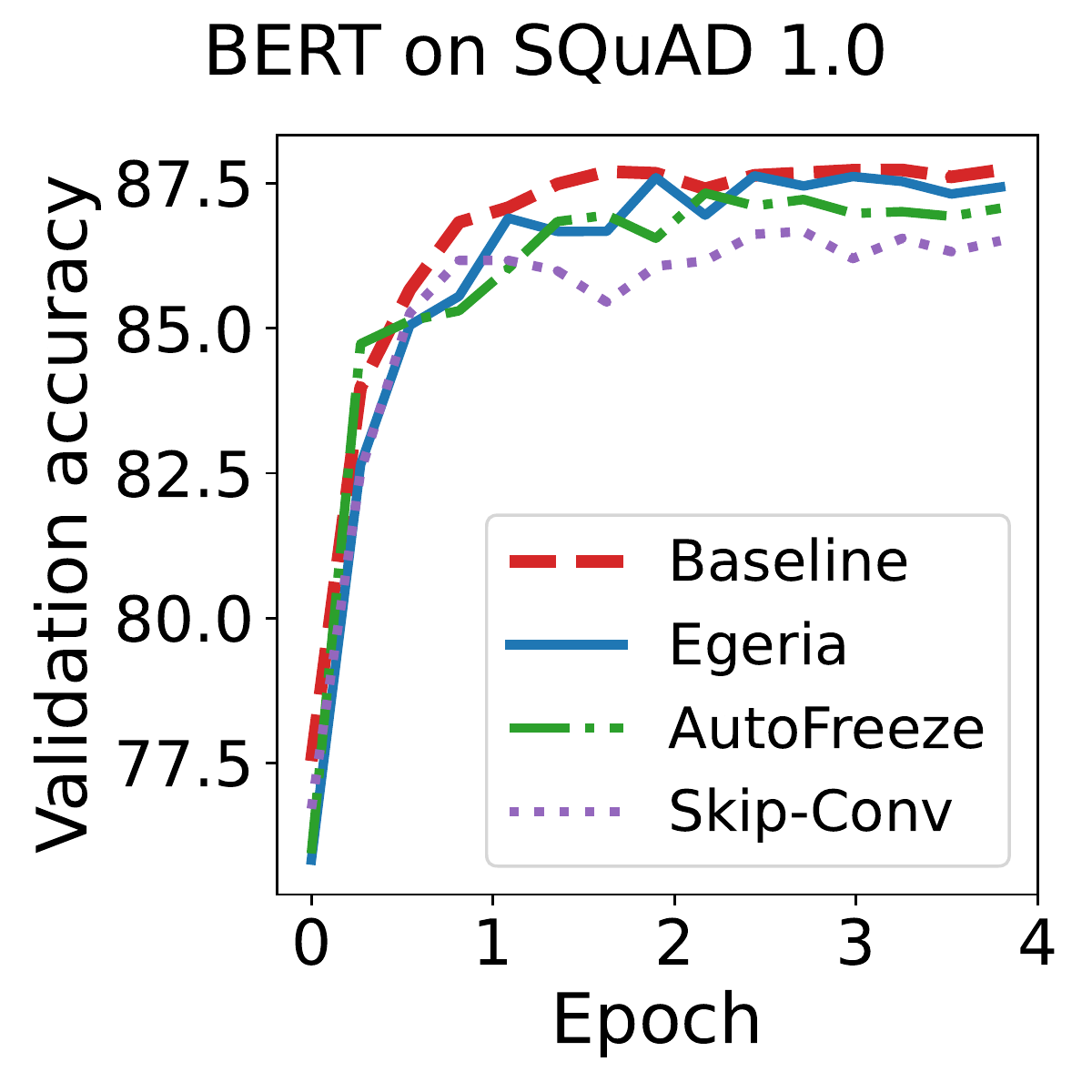}
      \caption{Fine-tuning speedups by 41\% while smaller accuracy gap.}
      \label{fig:qa}
    \end{subfigure}
    \caption{\name can accelerate training for different tasks without sacrificing accuracy compared to the full training baseline, while previous freezing techniques suffer from accuracy loss when reaching the same speedup (except for fine-tuning).}
    \label{fig:epoch-to-accuracy}
\end{figure*}

We use \name to train different models to reach the target accuracies with largely reduced training time.
Table~\ref{table:eval_summary} summarizes the evaluation results and the time-to-accuracy (TTA) speedups compared to the baseline training system.
We tune AutoFreeze and Skip-Conv to achieve a similar training time as \sys and compare their final accuracy to \sys and full training in Figure~\ref{fig:epoch-to-accuracy}.

\paragraph{Image classification.}

ResNet-50 for ImageNet is a popular CNN benchmarking model.
It consists of 48 layer modules, grouped into four stages, and the deep stages of layers have more parameters than the front stages (similar to the ResNet-56 structure in Figure~\ref{fig:cifar10_result_hist}).
Figure~\ref{fig:resnet50} shows the validation accuracy curves of \name and the baselines.
Within 90 epochs of training, \sys reaches the target accuracy while AutoFreeze and Skip-Conv each loses 1.5\% and 2.6\% when achieving the same time speedup of 28\%.
During these critical stages, the unfreezing mechanism of \name (\S\ref{subsubsec:howtounfreeze}) restarts the frozen layers and achieves the same level of accuracy boost.
The performance improvement primarily comes from later training stages when \name freezes the deeper layer modules with more parameters.
\name can accelerate lightweight models (e.g., MobileNet V2) on smaller datasets (e.g., CIFAR-10) with 22\% and 23\% speedups.

\paragraph{Semantic segmentation.}
We use the DeepLabv3 model with a ResNet-50 backbone for semantic segmentation training.
The structure of DeepLabV3 includes a backbone module for feature computation and extraction plus a classifier module that takes the output of the backbone and returns a dense prediction.
DeepLabv3 uses a Lambda LR scheduler that changes along with the training procedure, which will trigger the unfreezing mechanism of \name at the 45th epoch.
Figure~\ref{fig:deeplabv3} shows that, compared to the full baseline, \name can reach the target accuracy (mIoU of 63.3\%) 21\% faster and quickly improve accuracy at the later training stage when the LR scheduler significantly decreases; while the other freezing baselines lose accuracy by 2.1\% and 3\%.

\paragraph{Machine translation.}

\name not only works for CV models but also for language models.
A low perplexity means high accuracy for translation tasks.
In Figure~\ref{fig:nmt}, the model quickly reaches a low level of perplexity then continues to improve slowly.
\name brings a 43\% speedup by freezing the front encoders.
Unlike CNN models that usually have heavy deep layers, Transformer has a balanced structure, so skipping front layers can bring a considerable speedup.
The other freezing baselines each loses perplexity by 0.3 and 0.62.
We also evaluate Transformer-Tiny using an 8-GPU machine and achieve a 19\% speedup.

\paragraph{Question answering.}

Training a question answering model is different from the other tasks because we fine-tune a pre-trained general-purpose language model BERT for a new task on a new dataset, rather than training from scratch~\cite{devlin2018bert}.
Fine-tuning a pre-trained language model (e.g., BERT~\cite{devlin2018bert} and GPT-2~\cite{radford2019language}) is a popular training technique for NLP tasks because it can save computation overhead and achieve state-of-the-art results for many tasks, e.g., sequence classification and sentiment analysis.
The freezing technique was also first used in fine-tuning/transfer learning (see \S\ref{sec:related}).
Figure~\ref{fig:qa} shows the results of fine-tuning BERT on the SQuAD 1.0 dataset.
The metric for question answering is the F1 score.
\name accelerates the baseline by 41\% to reach the target accuracy, while AutoFreeze achieves a close performance compared to \sys as it is design for fine-tuning language models.
Since fine-tuning converges faster than training from scratch, \name does not freeze many deep layers before achieving the target, but the frozen front layers can still provide a good speedup.
During the training, the learning rate scheduler does not trigger the unfreezing mechanism.

\paragraph{Compared to freezing alternatives.}
\sys is motivated by the accuracy loss of training from scratch with freezing techniques designed for transfer learning, which is evaluated in Figure~\ref{fig:epoch-to-accuracy}.
AutoFreeze performs well in fine-tuning BERT but loses non-negligible final accuracy in other tasks, while \sys achieves the target accuracy in all tasks.
Using the metric of Skip-Conv loses more accuracy, which was designed for identifying differences in consecutive frames~\cite{habibian2021skip}.
When comparing models' intermediate results, Skip-Conv metric works similarly to an early KD research, FitNets~\cite{romero2014fitnets}, by directly subtracting two tensors.
Recent ML research suggests that, compared to such loss metric, SP loss can better capture the high-level similarity between activations~\cite{Tung_2019_ICCV}.

\subsection{Performance Breakdown}
\label{subsec:performance_breakdown}

\begin{figure}[!t]
    \includegraphics[width=\columnwidth]{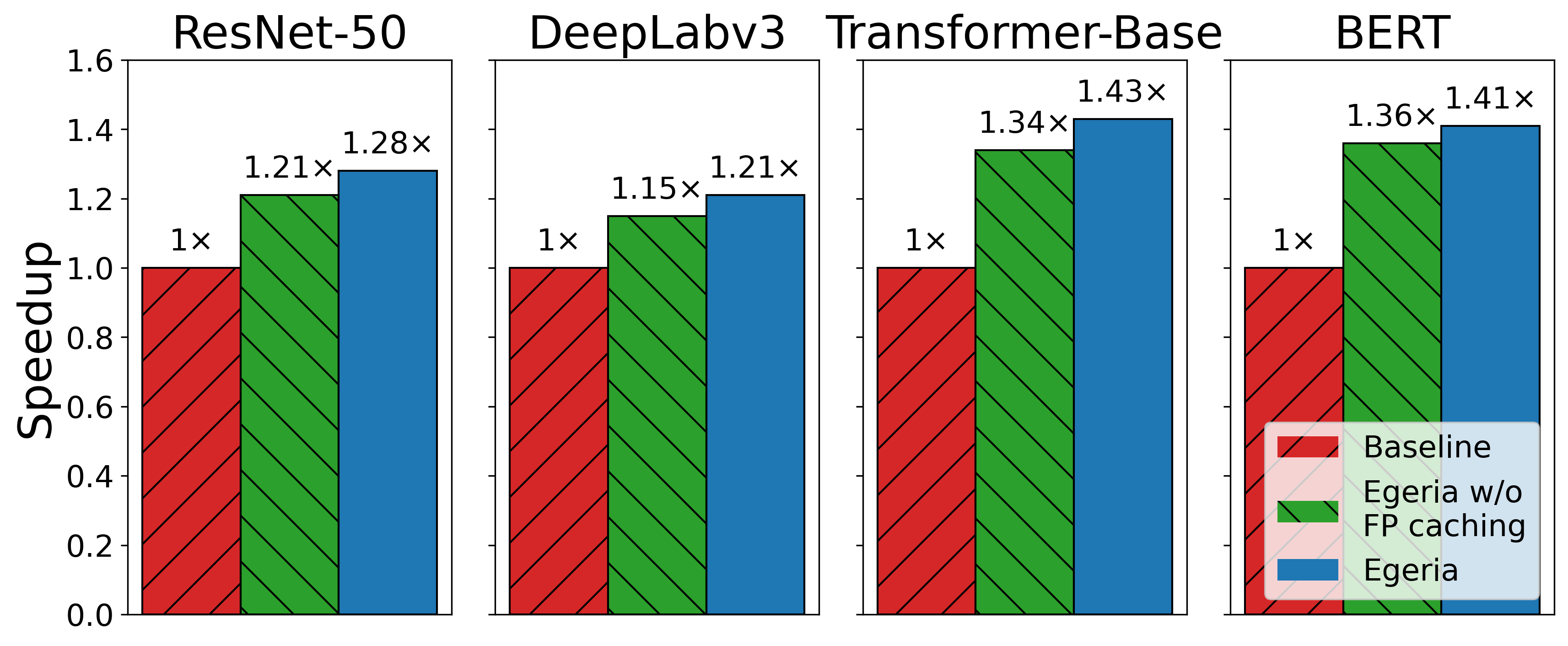}
        \caption{Performance breakdown of using layer freezing (middle) and prefetching FP pass (right).}
    \label{fig:breakdown}
\end{figure}

\paragraph{FP caching benefits.}
For single-node training, the performance speedup comes from the BP computation of the frozen layers and prefetching the cached FP results.
Figure~\ref{fig:breakdown} shows that caching FP generally contributes more for CNN models than language models but are all less than 10\%.
If there are few frozen layers or the front layers have fewer parameters, FP caching will be disabled.

\paragraph{Distributed training.}

\begin{figure}[!t]
  \includegraphics[width=\columnwidth]{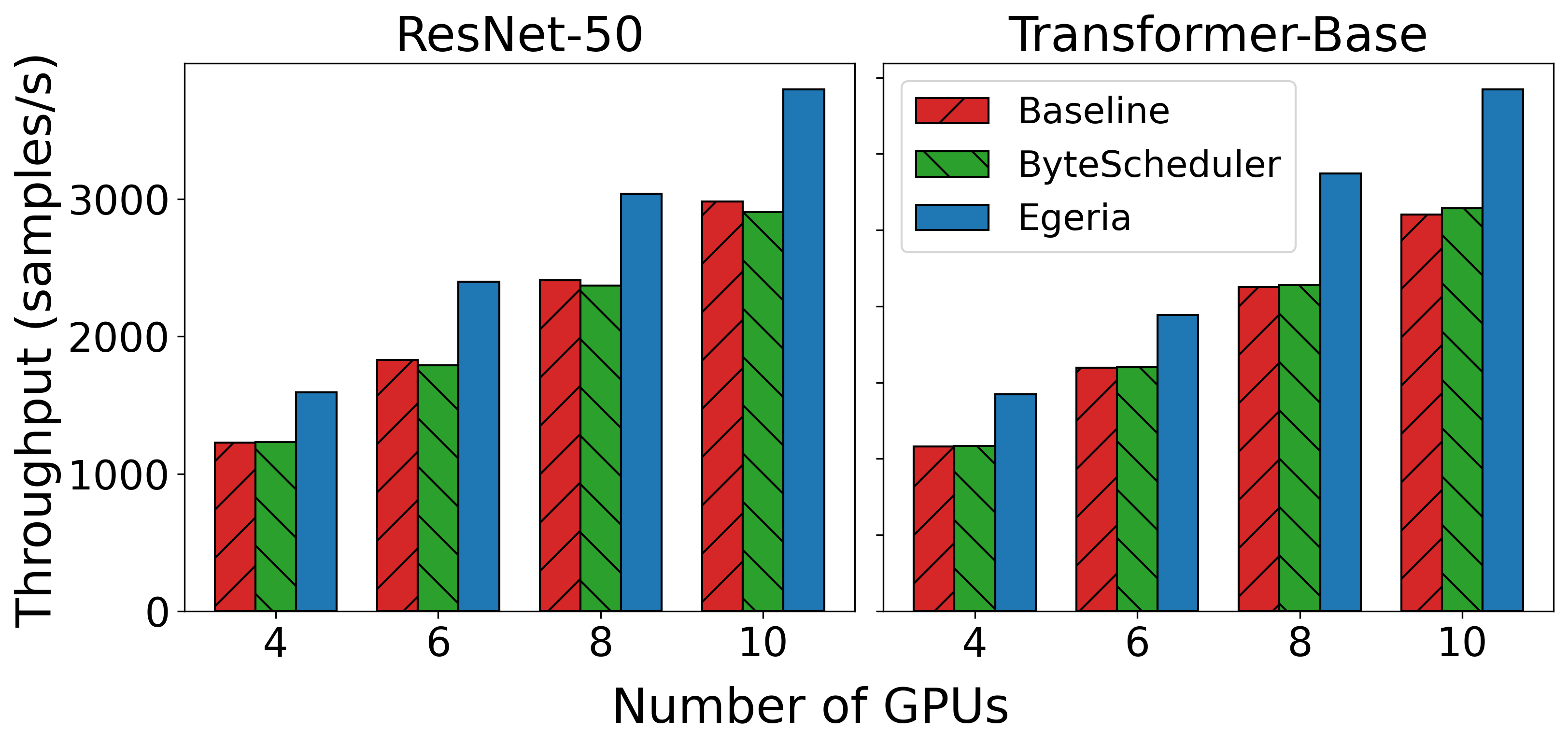}
  \caption{Distributed training performance. \name freezes layers to exclude them from parameter synchronization.}
  \label{fig:dist_throughput}
\end{figure}

\name accelerates multi-node data parallelism training as shown in Figure~\ref{fig:dist_throughput}.
\name can also work together with communication optimizations like ByteScheduler~\cite{peng2019generic}.
Other distributed methods, e.g., pipeline parallelism, can be explored in the future.

ResNet-50 and Transformer are both computation-intensive models, just like most of the recent DNN architectures, so the performance improvement of ByteScheduler is limited here.
A slight throughput drop when communication is not the bottleneck is normal for ByteScheduler with the default configuration~\cite{bs_issue}.
While the benefits of \name mostly come from computation saving, since frozen layers are not required for parameter synchronization, the reduced communication traffic can speedup the training by up to 5\% for ResNet-50, which can benefit the linear scalability of large scale training.

\begin{figure}[!t]
  \includegraphics[width=\columnwidth]{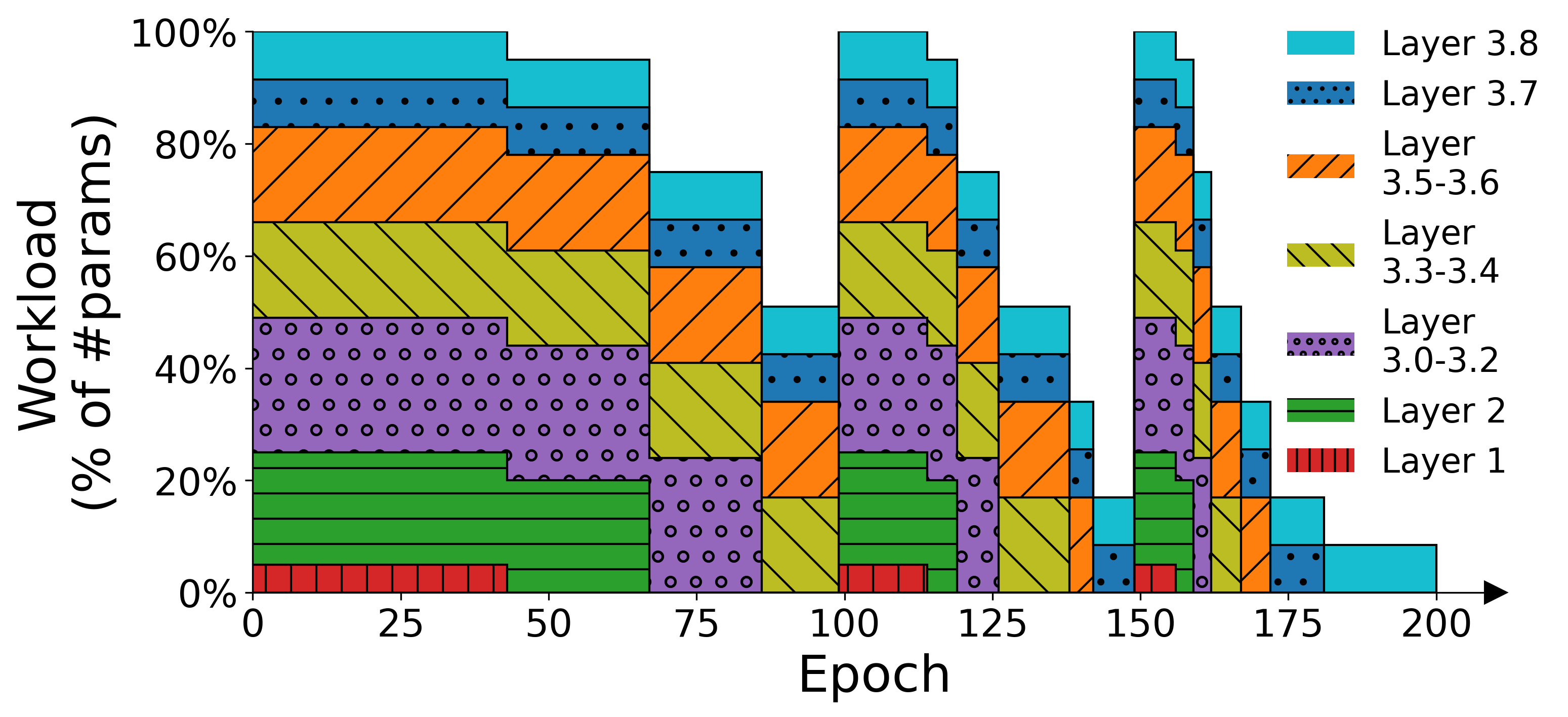}
  \caption{Freezing and unfreezing decisions breakdown through the ResNet-56's 200-epoch training. Y-axis shows the percentage of the active layers' parameters (their sizes).}
  \label{fig:cifar10_result_hist}
\end{figure}

\paragraph{Freezing \& unfreezing decisions.}
We take a closer look at one of our evaluations, training ResNet-56, to understand the decisions made by \name in Figure~\ref{fig:cifar10_result_hist}.
The bottom-up DNN consists of layer 1.0\textendash1.8, 2.0\textendash2.8, 3.0\textendash3.8, and input/output layers adjacent to layer 1.0 and 3.8.
\name parses the model based on its structure and the size of each layer, so that layer~3 (75\% of the total parameters), which is significantly larger than layer 2 (20\%), is split finer-grained into similar-sized modules; while layer 1 (5\%) and layer 2 are evaluated as a whole.
Layer 3.7\textendash3.8 (17\%) is further split because it is the last module.
\name gradually freezes layers and remarkably reduces the training cost (the blanks) without hurting accuracy.
Refreezing after the 100th and 150th epochs' unfreezing takes much less time because of the relaxed criteria (\S\ref{subsubsec:howtounfreeze}).

\subsection{Sensitivity Analysis}
\label{subsec:sensitivity}

\paragraph{Impact of the reference model's precision on accuracy.}

\name generates the reference model using int8 quantization by default for CPU execution efficiency (\S\ref{subsec:ref_model}).
We evaluate using higher precisions for the reference model, including float16 and float32 (full-precision), in ResNet-56 training on CIFAR-10, as shown in Table~\ref{table:precisions}.
We find using the int8-quantized reference model, which averagely has a 0.6\% lower accuracy, will not degrade the final accuracy and can largely improve the inference speed to obtain the intermediate activation.
Besides, \name can switch to higher-precision if int8 fails. 
Other tasks show similar results.

\begin{table}[!t]
    \centering
    \begin{tabular}{llll}
    \toprule
    Performance & int8 & float16 & float32 \\
    \midrule
    Final accuracy    & 92.1\% & 92.0\% & 92.2\% \\
    CPU inference speed & 3.59$\times$ & 1.69$\times$ & 1$\times$ \\
    Reference acc. gap  & -0.6\% & -0.2\% & 0 \\
    \bottomrule
    \end{tabular}
\caption{Using difference precisions for the reference model. \sys hits the sweet spot between efficiency and accuracy.}
\label{table:precisions}
\end{table}

\paragraph{Impact of hyperparameters on performance.}

\begin{figure}[!t]
    \centering
    \includegraphics[width=0.9\columnwidth]{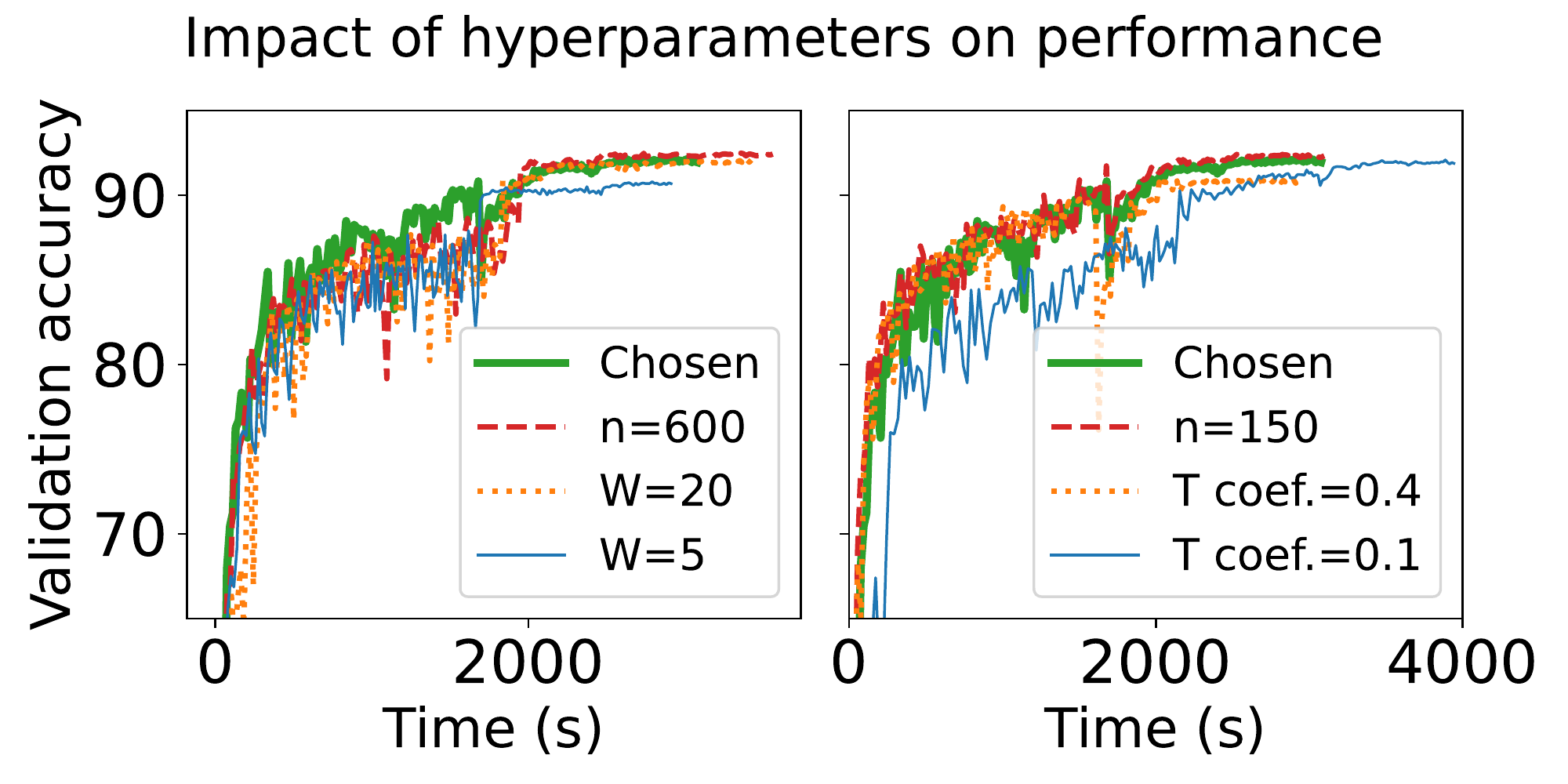}
    \caption{Following our hyperparameter guideline can balance accuracy and speedup (ResNet-56).}
    \label{fig:window_parameter}
\end{figure}

\sys evaluates plasticity in every $n$ iterations and uses the slope of linear fitting on a moving window $W$ to filter out the drastic fluctuation and provide a recent context.
If the plasticity slope has been considerably lower compared to itself during the early fast training stage (i.e., $s<T$) for $W$ evaluations, we freeze the layer.
We find that the hyperparameters are tolerant in general when following our guidelines, while drastically changing them could result in performance penalties.
As shown in Figure~\ref{fig:window_parameter}, halving $W$ from 10 to 5 or doubling $T$'s coefficient from 20\% to 40\% would eagerly freeze unconverged layers, hurt the accuracy, but only make training slightly faster, while doubling $W$ to 20 or evaluation interval $n$ from 300 to 600 would lead to longer training time without accuracy gain.
Halving $T$'s coefficient to 10\% virtually disables freezing.
Making frequent evaluations ($n$=150) brings no extra speedup, while further reducing $n$ could consume more CPUs and potentially slow down the training.

\subsection{System Overhead and Discussion}
\label{subsec:overhead}

\name leverages CPUs to freeze layers accurately while maintaining accuracy; it also uses disk storage to reduce forward computation overhead.
As a result, \name reduces the training time by 19\%-43\%.
Through careful system designs, we minimize the extra overhead of \name.

\paragraph{The reference model.}
Using the reference model for plasticity evaluation involves the model generation and execution.
We find that generating and updating the reference model through dynamic and static quantization on CPU take 0.5s\textendash1.5s for each time, thus bring no noticeable slowdown.
Running the reference model on CPU could introduce up to 1.5\% time overhead to the overall training process on standard training server configurations, which is worthwhile compared to the saving from layer freezing.
If CPU resources are limited (e.g., on shared machines or using CPU-based optimizations), we support GPU execution for the reference model.

\paragraph{Caching and prefetching.}
We store the serialized intermediate tensors of the frozen layers to the disk for prefetching.
The storage usage depends on the DNN architecture.
For example, we need 1.5$\times$ to 5.3$\times$ compared to the input for ResNet-50, which is generally viable.
For language models, since the text data volume is smaller, the overall space usage is limited.
Since we only keep the relevant tensors in memory, the overhead is small compared to the regular utilization.
It takes hundreds of MB of GPU memory, which is a small fraction of device memory for modern GPUs.

\paragraph{Generalization.}
We design \name as a general system.
Users can adjust the usage of CPU and storage in plasticity evaluation and caching to meet their needs.
Future research can study how \name collaborates with other CPU-based (e.g., BytePS~\cite{jiang2020unified}) or storage-based (e.g., CoorDL~\cite{coordl}) optimizations and on different hardware.
\section{Related Work}
\label{sec:related}
\paragraph{Efficient training systems.}

Accelerating DNN training is a key goal of ML systems.
To optimize the computation, they may optimize the computation graph to maximize the degree of parallelism~\cite{taso}, or deploy advanced scheduling to distribute the computation across multiple machines~\cite{tiresia, themis}.
To maximize the communication efficiency, priority-based communication scheduling systems~\cite{peng2019generic,jayarajan2019priority} use the layered structural information to prioritize the front layers and avoid blocking layers with high priority.
BytePS~\cite{jiang2020unified} combines the benefits of parameter server and all-reduce and transmits the gradients among workers or between workers and servers.
Some efforts~\cite{lin2018deep,fetchsgd} measure the importance of gradient updates in terms of the magnitude of difference w.r.t. the last update, and then filter out trivial parameters before shipping the updates.
\name aims to reduce the total training workload, thus should be compatible with them.
Additionally, there are a wide range of networking solutions that can help in distributed DNN training~\cite{ma2022autobyte,wang2022addressing,zeng2022herald,wan2022dgs,wang2020domain}.
ModelKeeper~\cite{modelkeeper-nsdi23} accelerates training by repurposing previously-trained models in a shared cluster.
Oort~\cite{lai2021oort,lai2022fedscale} accelerates federated training with guided participant selection.

\paragraph{Using an assistant model in training.}

\sys echoes the broad idea of using another DNN to assist training.
Knowledge distillation trains a small student model to mimic the probability distribution of a pre-trained large model~\cite{hinton2015distilling}.
Co-distillation~\cite{anil2018large} trains multiple tweaked copies of the model in a distributed manner and encourages one model to agree with others' predictions.
AutoAssistant~\cite{NEURIPS2019_9bd5ee6f} trains a lightweight assistant model to identify the hard-to-classify examples and feed them to the training model to improve its performance fast.
Infer2Train~\cite{hoffer2018infer2train} runs a copy of the training model on the additional hardware accelerator and finds the difficult examples to prioritize in the following iterations.

\paragraph{Freezing parameters and caching DNN results.}
Existing proposals on freezing are limited to fine-tuning certain models~\cite{liu2021autofreeze,pmlr-v139-he21a,Guo_2019_CVPR} or reducing communication only~\cite{chen2021communication}; otherwise, considerable accuracy loss would nullify any improvements in training speed~\cite{brock2017freezeout,lee2019would}.
An early work FreezeOut~\cite{brock2017freezeout} explores the freezing technique in general training with heuristics but reports large accuracy loss on many models; nevertheless, it shows that freezing can trade off accuracy for speed.
A concurrent work AutoFreeze~\cite{liu2021autofreeze} focuses on fine-tuning pre-trained Transformer-based models; it falls into the original use of transfer learning rather than general training and we found that fine-tuning suffers less from accuracy loss than training from scratch (discussed in \S\ref{subsec:oppo} and \S\ref{subsec:end2end}).
PipeTransformer~\cite{pmlr-v139-he21a} also applies freezing in fine-tuning Transformers with pipeline parallelism using a gradient-based importance metric~\cite{xiao2019fast}; still, it novelly explores opportunities in pipeline parallelism.
We discuss the accuracy performance of gradient-based metrics in \S\ref{subsubsec:plasticity}.
APF~\cite{chen2021communication} excludes stable parameters from synchronization in federated learning; it suggests that model snapshots can best capture the performance and implements a workaround.
GATI~\cite{balasubramanian2021accelerating} accelerates DNN inference by caching the intermediate results and skipping the rest of the forward pass.

\section{Conclusion}
\label{sec:conclusion}

We introduce a novel system \name to accelerate DNN training while maintaining accuracy by accurately freezing the converged layers.
To avoid the limitations of existing work, we employ a reference model and use semantic knowledge to evaluate the \emph{plasticity} of internal layers efficiently during training.
\name excludes the frozen layers from the backward pass and parameter synchronization.
Furthermore, we cache the frozen layers' intermediate computation with prefetching to skip the forward pass.
We evaluate \name using several CV and language models and find that \name can accelerate training by 19\%-43\% without hurting accuracy.

\section*{Acknowledgment}

We thank the anonymous EuroSys reviewers and our shepherd Dr. Jayashree Mohan for their constructive feedback and suggestions.
This work is supported in part by the Key-Area R\&D Program of Guangdong Province (2021B0101400001), the Hong Kong RGC TRS T41-603/20-R, GRF-16213621, ITF-ACCESS, the NSFC Grant 62062005, and the Turing AI Computing Cloud (TACC)~\cite{tacc}. 
Fan Lai and Mosharaf Chowdhury were partly supported by National Science Foundation grants (CNS-1900665, CNS-1909067, CNS-2106184).
Kai Chen is the corresponding author.

\bibliographystyle{plain}
\bibliography{reference}

\end{document}